\documentclass{article}

\usepackage[final]{corl_2026} 

\usepackage[utf8]{inputenc}
\usepackage[T1]{fontenc}
\usepackage{url}
\usepackage{booktabs}
\usepackage{amsfonts}
\usepackage{nicefrac}
\usepackage{microtype}
\usepackage{xcolor}
\usepackage{amsmath,amssymb,multirow,graphicx,enumitem}
\usepackage{tikz}
\usepackage{pgfplots}
\usepackage{todonotes}
\pgfplotsset{compat=1.18}
\usetikzlibrary{arrows.meta,decorations.pathreplacing,positioning}
\title{Beyond Ensemble Disagreement: Predictive STL Monitoring of RL Locomotion under Distribution Shift}

\author{%
  Qurban Ali$^{1}$, Badaruddin Chachar$^{2}$, Maqsood Ali$^{3,4}$, Sajjad Ali Rajper$^{5}$\\
  1. Sindh Madressatul Islam University, Karachi, Pakistan\\
  2. Shah Abdul Latif University, Ghotki, Pakistan\\
  3. University of Camerino, Camerino, Italy\\
  4. Institute of Biophysics, CNR, Palermo, Italy\\
  5. University of Mirpurkhas, Mirpurkhas, Pakistan\\
  \texttt{qurban@smiu.edu.pk$^{1}$}; \texttt{badaruddin.chachar@salu.edu.pk$^{2}$}\\
  \texttt{maqsoodali@cnr.it$^{3,4}$}; \texttt{sajjadali@umpk.edu.pk$^{5}$}
}
\begin{document}\maketitle
\begin{abstract}
RL locomotion policies trained under nominal simulation conditions may encounter distribution shifts that compromise safety. Predictive monitors could anticipate violations, but their world-model forecasts may become unreliable under shift, and ensemble disagreement is not calibrated. We evaluate predictive Signal Temporal Logic (STL) monitoring for frozen Ant-v5 and HalfCheetah-v5 policies using closed-loop world-model rollouts and a three-model ensemble. Across external-force, observation-noise, and mass shifts, predicted robustness ranks HalfCheetah's future STL violations strongly, whereas disagreement does not reliably track observation prediction error or violations. Under severe noise, violations remain common ($64.2\pm24.4\%$ at $K{=}20$ across three PPO seeds), yet current-state robustness without prediction comes within $0.03$ AUROC of predictive robustness. Penalizing predicted robustness by disagreement does not consistently improve ranking. Ant provides complementary per-specification evidence, but its composite target has no positive violations and its any-violation target is single-class in most evaluated seed-condition slices. Overall, predictive robustness provides useful STL-specific ranking but little demonstrated additional ranking value over the current state; these findings caution against treating ensemble disagreement as a reliable indicator of safety uncertainty.
\end{abstract}

\section{Introduction}
\label{sec:introduction}

Reinforcement learning (RL) has become a powerful approach for learning locomotion policies in simulated robotic environments~\cite{ppo}. However, policies trained under fixed simulation conditions often encounter distribution shifts during deployment, such as external disturbances, observation noise, changes in physical parameters, or variations in terrain properties, which can compromise safety. These shifts can cause degraded locomotion, loss of balance, or entry into states that violate safety requirements.

To address this challenge, runtime safety monitoring complements policy training: a monitor evaluates a frozen policy against explicit safety specifications during deployment. Signal Temporal Logic (STL) suits this purpose because quantitative robustness values continuously measure satisfaction or violation~\cite{stl}. Runtime monitors can be broadly classified as reactive or predictive. Reactive monitors assess observed states after the fact---including interpretable STL-based retrospective monitoring~\cite{donze2011robust,yamaguchi2024rtamt} and simpler threshold- or barrier-style checks---whereas predictive monitors anticipate future violations. Learned world models enable predictive monitoring by rolling the policy forward in closed loop and evaluating safety specifications on forecasted states before they are reached~\cite{dreamer} (distinct from model-based RL planning and learning). We instantiate the world model as a lightweight feedforward latent-dynamics ensemble (not a recurrent/stochastic Dreamer-class model), so conclusions about disagreement informativeness are scoped to this architecture class (see Sec.~\ref{sec:discussion}). A key challenge is that prediction quality may degrade under distribution shift, so a monitor that ignores model uncertainty may make confident but incorrect safety assessments.


Deep ensembles provide a practical measure of predictive disagreement
\citep{ensemble}, but whether disagreement reliably indicates
prediction error or future safety violations under distribution
shift remains unclear. In particular, disagreement is not equivalent
to calibrated predictive uncertainty or safety risk. Three gaps motivate this study: (i) predictive STL monitoring has
received less attention than retrospective monitoring
\cite{donze2011robust,yamaguchi2024rtamt,predictive_safety,
meng2023stlnpc,han2025mpmonitoring};
(ii) the reliability of ensemble disagreement under distribution
shift remains insufficiently understood~\cite{ensemble, zhao2024conformalstl, cobbe2020quantifying}; and
(iii) its usefulness for downstream safety-failure prediction has
not been systematically compared across observation, latent, and
STL representations using failure-prediction and selective
prediction metrics~\cite{cairoli2025conformal,predictive_safety,beirami2026conformal}.


To address these gaps, this paper asks: \emph{when a learned world model predicts that a future trajectory is safe, can ensemble disagreement identify situations in which that prediction is unreliable under distribution shift?} Let $\hat{\rho}$ denote predicted STL robustness and $\rho$ the robustness of the corresponding observed trajectory. We empirically examine (H1) $U_\rho \uparrow \Rightarrow |\hat{\rho}-\rho| \uparrow$ and (H2) $U_\rho \uparrow \Rightarrow \Pr(Y_t^{(K)}{=}1) \uparrow$. We develop and evaluate a predictive monitoring framework on Ant-v5~\cite{antv5} and HalfCheetah-v5~\cite{halfcheetahv5} under external forces, observation noise, and mass variation. We report quantitative window-level failure-prediction results for HalfCheetah-v5; Ant-v5 serves primarily as a per-specification complement (Sec.~\ref{sec:extension}). Across the evaluated settings, we find no consistent evidence that disagreement reliably predicts
error or future violations. Predicted robustness ranks failures
more strongly, but a current-state robustness baseline recovers most of
that ranking, so the evidence for predictive (as opposed to reactive)
monitoring is modest.

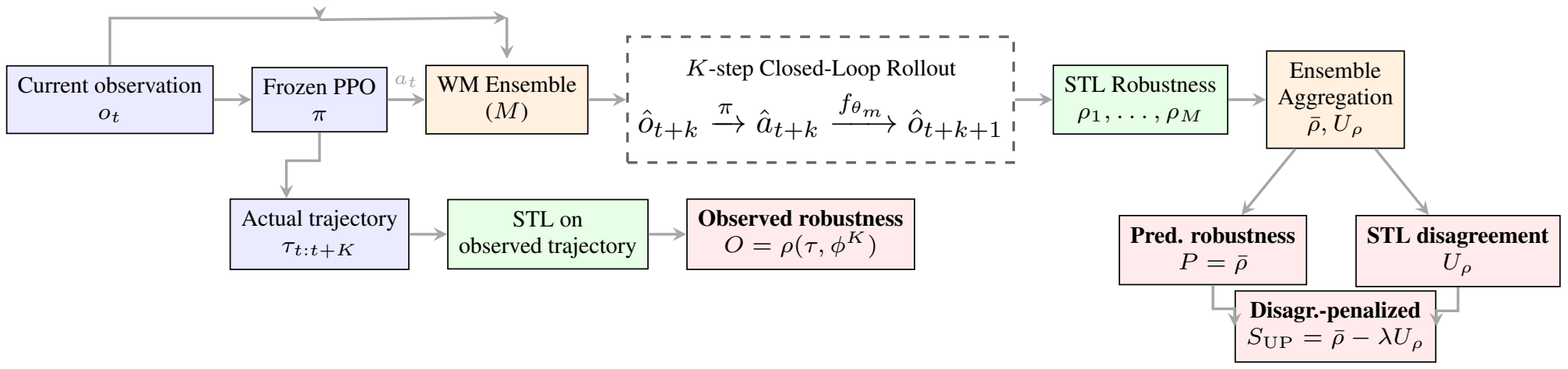
\begin{figure*}[t]
\centering
\resizebox{0.95\textwidth}{!}{%
\begin{tikzpicture}[
  >=stealth,
  block/.style={
    rectangle, draw, fill=blue!8,
    minimum height=1.6em, minimum width=3.2em,
    align=center, font=\scriptsize
  },
  model/.style={
    rectangle, draw, fill=orange!12,
    minimum height=1.6em, minimum width=3.2em,
    align=center, font=\scriptsize
  },
  stl/.style={
    rectangle, draw, fill=green!10,
    minimum height=1.6em, minimum width=3.2em,
    align=center, font=\scriptsize
  },
  output/.style={
    rectangle, draw, fill=red!8,
    minimum height=1.6em, minimum width=3.5em,
    align=center, font=\scriptsize
  },
  arr/.style={->, thick, gray!70},
  dashedbox/.style={
    rectangle, draw=black!60, dashed, thick,
    minimum height=3.8em, minimum width=7.0em,
    align=center, inner sep=3pt
  },
]

\node[block] (obs) {Current observation\\$o_t$};

\node[block, right=0.4cm of obs] (ppo)
  {Frozen PPO\\$\pi$};

\node[model, right=0.4cm of ppo] (wm)
  {WM Ensemble\\($M$)};

\node[dashedbox, right=0.4cm of wm] (rollout) {
  {\scriptsize $K$-step Closed-Loop Rollout}\\[2pt]
  $\hat{o}_{t+k}
  \xrightarrow{\pi}
  \hat{a}_{t+k}
  \xrightarrow{f_{\theta_m}}
  \hat{o}_{t+k+1}$
};

\node[stl, right=0.4cm of rollout] (stl_pred)
  {STL Robustness\\$\rho_1,\ldots,\rho_M$};

\node[model, right=0.4cm of stl_pred] (agg)
  {Ensemble\\Aggregation\\$\bar{\rho},U_\rho$};

\node[block, below=0.7cm of ppo] (traj)
  {Actual trajectory\\$\tau_{t:t+K}$};

\node[stl, right=0.4cm of traj] (stl_obs)
  {STL on\\observed trajectory};

\node[output, right=0.4cm of stl_obs] (O) {
  \textbf{Observed robustness}\\
  $O=\rho(\tau,\phi^K)$
};

\node[output, below=0.7cm of agg, xshift=-1.3cm] (P) {
  \textbf{Pred. robustness}\\
  $P=\bar{\rho}$
};

\node[output, below=0.7cm of agg, xshift=1.3cm] (U) {
  \textbf{STL disagreement}\\
  $U_\rho$
};

\node[output, below=1.5cm of agg] (SUP) {
  \textbf{Disagr.-penalized}\\
  $S_{\mathrm{UP}}=\bar{\rho}-\lambda U_\rho$
};

\draw[arr] (obs) -- (ppo);
\draw[arr] (obs.north) -- ++(0,0.5) -| ([yshift=0.5cm]ppo.north);
\draw[arr] ([yshift=0.5cm]ppo.north) -- ([yshift=0.5cm]wm.north);
\draw[arr] ([yshift=0.5cm]wm.north) -- ([yshift=0.4ex]wm.north);

\draw[arr] (ppo) -- node[font=\tiny, above] {$a_t$} (wm);

\draw[arr] (wm) -- (rollout);
\draw[arr] (rollout) -- (stl_pred);
\draw[arr] (stl_pred) -- (agg);

\draw[arr] (ppo.south) -- ++(0,-0.3) -| ([xshift=-1em]traj.north);
\draw[arr] (traj) -- (stl_obs);
\draw[arr] (stl_obs) -- (O);

\draw[arr] (agg) -- (P);
\draw[arr] (agg) -- (U);

\draw[arr] (P.south) -- ++(0,-0.25) -| (SUP.west);
\draw[arr] (U.south) -- ++(0,-0.25) -| (SUP.east);

\end{tikzpicture} \vspace{-8pt}%
}

\caption{
Predictive STL monitoring pipeline.
The current observation $o_t$ feeds the frozen PPO policy and the world-model ensemble, whose members each generate a $K$-step closed-loop rollout by repeatedly querying the policy on predicted observations.
Predicted trajectories are evaluated with STL and aggregated into predictive robustness $P$ and STL disagreement $U_\rho$; independently, $O$ is evaluated on the observed trajectory $\tau_{t:t+K}$, and $S_{\mathrm{UP}}$ penalizes $P$ by $U_\rho$.
}
\vspace{-8pt}
\label{fig:framework}
\end{figure*}

The contributions of this paper are as follows:
\begin{enumerate}[leftmargin=*]

\item \textbf{Framework.} We develop a predictive STL monitoring framework for frozen RL locomotion policies based on closed-loop, policy-conditioned world-model rollouts, using ensemble variation in predicted STL robustness as an empirical disagreement measure (not a calibrated estimate; Section~\ref{sec:disagreement}).

\item \textbf{Evaluation.} We evaluate on Ant-v5 and HalfCheetah-v5 using independently trained PPO policies and a primary multi-seed evaluation over three PPO seeds. We compare retrospective and predictive STL monitoring under external forces, observation noise, and mass variation, with failure-prediction, association, and horizon analyses. We add a current-state robustness baseline and a safe-at-start analysis
that separates early warning from detection of already-unsafe states.

\item \textbf{Replication.} To support reproducibility, we provide the replication package at~\cite{replication}.

\end{enumerate}

\section{Related Work}
\label{sec:related}

Signal Temporal Logic (STL) provides quantitative robustness
semantics for specifying and monitoring temporal requirements in
cyber-physical and robotic systems~\cite{stl}. Runtime monitoring
frameworks such as RTAMT~\cite{yamaguchi2024rtamt} build on
foundational online monitoring approaches~\cite{donze2011robust}.
Predictive runtime verification extends this setting by evaluating
predicted future trajectories, including uncertainty-aware
monitoring~\citep{predictive_safety}, neural predictive control
\citep{meng2023stlnpc}, and nested-STL monitoring
\citep{han2025mpmonitoring}. Conformal approaches further investigate
calibrated or guaranteed prediction under different distributional
assumptions~\citep{cairoli2025conformal,zhao2024conformalstl,
beirami2026conformal}.

Our work uses learned world models to generate policy-conditioned
future trajectories and deep ensembles to measure variation in
predicted STL robustness~\citep{dreamer,ensemble}. Unlike approaches
that aim to provide calibrated uncertainty or formal guarantees,
we empirically investigate whether ensemble disagreement ranks
prediction errors and future safety violations under distribution
shift. We compare disagreement in observation, latent, and STL
representations on common downstream failure-prediction tasks.

\section{Problem Formulation}
\label{sec:problem}

We consider a frozen reinforcement learning policy deployed in an environment that may differ from the training conditions. The policy maps observations to actions but has no explicit mechanism for detecting or adapting to distribution shift, which motivates the need for an independent safety monitor that operates alongside the policy.

\subsection{Policy and Environment}

At time $t$, environment state $s_t$ produces observation $o_t=h(s_t)$,
the frozen policy selects action $a_t=\pi(o_t)$, and the state evolves as
$s_{t+1}=f(s_t,a_t,\xi_t)$. Here, $h$ is the observation function, $\pi$
is the policy, $f$ is the environment transition function, and $\xi_t$
captures stochasticity and unmodelled effects~\cite{ppo}. Observations may
also include noise or partial information. Under distribution shift, the
effective dynamics differ from those seen during training, while the policy
continues to act as if nominal conditions hold.

\subsection{Learned World Model}

To predict whether future trajectories will satisfy safety specifications, we employ a learned world model that approximates the environment dynamics in observation space~\cite{dreamer}. The world model consists of three components: an encoder $e_\theta$ that maps observations to a latent state, a latent dynamics model $d_\theta$ that predicts the next latent state given the current latent state and action, and a decoder $g_\theta$ that reconstructs observations from the latent state:
\begin{align}
z_t&=e_\theta(o_t),\\
\hat{z}_{t+1}&=d_\theta(z_t,a_t),\\
\hat{o}_{t+1}&=g_\theta(\hat{z}_{t+1}).
\end{align}
where $o_t$ is the observation at time $t$, $z_t$ its encoded latent state,
$\hat{z}_{t+1}$ the predicted next latent state given action $a_t$,
$\hat{o}_{t+1}$ the decoded next-observation prediction, and $\theta$ denotes their learned parameters.
The complete observation-prediction model is
$f_\theta(o_t,a_t)=g_\theta(d_\theta(e_\theta(o_t),a_t))$.
We independently train $M$ copies of this model to form the
ensemble.

\subsection{Closed-Loop Predictive Rollout}

For predictive monitoring at time $t$, we perform a $K$-step closed-loop rollout in which the frozen policy acts on each latest predicted observation, rather than propagating the dynamics open-loop:
\begin{align}
\hat{a}_{t+k}&=\pi(\hat{o}_{t+k}),\\
\hat{o}_{t+k+1}&=g_\theta(d_\theta(e_\theta(\hat{o}_{t+k}),\hat{a}_{t+k})),\quad k=0,\ldots,K{-}1,
\end{align}
where $\hat{a}_{t+k}$ is the action selected by
the policy from predicted observation $\hat{o}_{t+k}$, and
$\hat{o}_{t+k+1}$ is the corresponding next-observation prediction. Feeding each predicted observation
back to the policy captures policy-induced feedback absent from open-loop
prediction. The safety-variable mapping $q(\cdot)$ used to extract STL signals is detailed in Appendix~\ref{app:sup-details}.

\section{STL Monitoring}
\label{sec:stl}

Signal Temporal Logic provides a formal framework for specifying temporal safety requirements over continuous-valued signals~\cite{stl,donze2011robust}. Each specification constrains the evolution of safety variables over a temporal window, and its quantitative robustness value indicates the degree of satisfaction (positive) or violation (negative). We use STL in two complementary modes: retrospective evaluation of observed trajectories and predictive evaluation of model-generated future trajectories. Appendix~\ref{app:sup-details} provides the complete set of episode-level specifications.

\subsection{Predictive Horizon Formulation}

For horizons $K\in\{1,5,10,20\}$, we evaluate the instantaneous
properties using truncated specifications
\begin{equation}
\phi_i^K=G_{[0,K]}(\mathrm{property}_i),
\quad i=1,\ldots,5.
\end{equation}
where $G_{[0,K]}$ is the bounded ``always'' operator over the inclusive
interval $[0,K]$, and $\mathrm{property}_i$ denotes the $i$th instantaneous
safety predicate. The interval contains $K+1$ observations connected by
$K$ transitions. The recovery property $\phi_6$ requires a 50-step
response interval and is therefore excluded from the predictive
composite at the evaluated horizons. Its truncated pastified
robustness is nevertheless retained in per-specification
diagnostics and the any-violation target. These truncated values
describe the available trace prefix, not satisfaction of the
complete recovery obligation (Appendix~\ref{app:sup-details}). 

The predictive composite score combines the remaining properties:
\begin{equation}
S_t^{P}(K)=\frac{\sum_{i\in\mathcal{I}_e(K)}w_i\rho_i^K(t)}
{\sum_{i\in\mathcal{I}_e(K)}w_i},
\end{equation}
where $\mathcal{I}_e(K)$ is the set of applicable specifications
for environment $e$ at horizon $K$, and $\rho_i^K(t)$ is the
robustness of property $i$ over the window starting at time $t$; $w_i$ is
the aggregation weight assigned to specification $i$.
For Ant-v5, five properties contribute; for HalfCheetah-v5,
pitch, height, and contact contribute, with contact providing
no within-environment discrimination.

Where we report a common-structure composite, it is secondary to per-specification binary violation prediction, the primary evaluation (Sec.~\ref{sec:results}). Specifications and the windowing stride ($q{=}\max(1,K)$, with boundary exclusions) are hand-authored design choices, and all reported experiments use equal weights. The tested height-threshold variation leaves the $-P{\gg}U_\rho$ ranking unchanged (Tab.~\ref{tab:height_sens}); sensitivity to alternative specifications, recovery thresholds, aggregation weights, strides, and unit-normalized composites remains untested.

\subsection{Retrospective versus Predictive Monitoring}

The retrospective observed-trajectory reference score evaluates the trajectory that actually occurred over the corresponding $K$-step window:
\begin{equation}S_t^{O}(K)=\rho(\tau_{t:t+K},\phi^K),\end{equation}
where $\tau_{t:t+K}$ is the observed trajectory, and $\rho(\cdot,\phi^K)$ returns
their composite robustness. Evaluating this score requires the full future
trajectory; it serves as a reference for predictive scores. We evaluate
model-generated trajectories through closed-loop policy rollouts and average
their robustness across ensemble members, yielding $P{=}\bar{\rho}$ by
construction:
\begin{equation}S_t^{P}(K)=\frac{1}{M}\sum_{m=1}^{M}\rho(\hat{\tau}^{(m)}_{t:t+K},\phi^K)=\bar{\rho}(t),\end{equation}
where $M$ is the ensemble size, $m$ indexes its members, and
$\hat{\tau}^{(m)}_{t:t+K}$ is member $m$'s predicted trajectory. The term
$\bar{\rho}(t)$ is the mean predicted robustness across members. We use the
difference between $S_t^{O}(K)$ and $S_t^{P}(K)$ to characterize predictive
bias in the world model.

\section{Ensemble Disagreement for World-Model Monitoring}
\label{sec:uncertainty}

We use independently trained world models to measure predictive disagreement and test whether this disagreement identifies unreliable safety predictions under distribution shift. We refer to the main measure as STL robustness disagreement rather than calibrated uncertainty.

\subsection{Ensemble Dynamics}

We train an ensemble of $M$ independently initialized world models
$\{f_{\theta_1},\ldots,f_{\theta_M}\}$~\cite{ensemble}, each using
the observation-prediction model defined in
Sec.~\ref{sec:problem}. Each model independently generates a
$K$-step closed-loop prediction using the rollout procedure
described in Sec.~\ref{sec:problem}. All models are trained on
nominal locomotion data collected under standard conditions and
are not adapted to shifted test trajectories. This allows us to
examine disagreement under distribution shift without introducing
test-time adaptation effects.

\subsection{Prediction Error and STL Robustness Disagreement}
\label{sec:disagreement}

We compare prediction error with disagreement in predicted STL
robustness. For each ensemble member, prediction error is the
mean $L_2$ distance over the $K$ predicted future observations;
the initial observation $o_t$ is the known rollout input and is
not included:
\begin{equation}
E_t^{(m)} =
\frac{1}{K}\sum_{k=1}^{K}
\|\hat{o}^{(m)}_{t+k}-o_{t+k}\|_2,
\end{equation}
where $m$ indexes an ensemble member, $\hat{o}^{(m)}_{t+k}$ is its predicted
observation, and $o_{t+k}$ is the corresponding observed future observation with ensemble mean error
$E_t=M^{-1}\sum_{m=1}^{M}E_t^{(m)}$. This provides an empirical reference for prediction accuracy,
rather than a direct measure of safety. Our primary disagreement measure is the standard deviation of
predicted STL robustness across ensemble members:
\begin{equation}
U_\rho(t)=
\sqrt{\frac{1}{M}\sum_{m=1}^{M}
(\rho_m(t)-\bar{\rho}(t))^2},
\end{equation}
where $\rho_m(t)$ is the predicted STL robustness from member
$m$, and $\bar{\rho}(t)=M^{-1}\sum_{m=1}^{M}\rho_m(t)$.

High $U_\rho$ indicates variation in predicted STL robustness,
whereas low disagreement does not imply correctness: ensemble
members may share a common prediction bias. Thus, $U_\rho$ is
an empirical disagreement measure, not calibrated uncertainty
or a safety-risk estimate. For comparison, we also evaluate observation-space and
latent-space disagreement, denoted by $U_o$ and $U_z$,
respectively. These measures quantify ensemble variation in
the corresponding representations and are computed on the
same predicted rollouts. Their precise definitions are given
in Appendix~\ref{app:sup-disagreement}.

\subsection{Disagreement-Penalized Robustness Score}

To examine whether disagreement can improve robustness-based
failure detection, we define the penalized score
\begin{equation}
S_{\mathrm{UP}}(t)=
\bar{\rho}(t)-\lambda U_\rho(t),
\end{equation}
where $\lambda\geq0$ controls the disagreement penalty.
In our implementation, $\bar{\rho}$ corresponds to the
ensemble-mean robustness score $P$, so that
$S_{\mathrm{UP}}=P$ when $\lambda=0$.
This is an empirical penalty, not a formal probabilistic
confidence bound. We evaluate
$\lambda\in\{0,1,2,5\}$ without tuning it on test conditions,
and flag a window when $S_{\mathrm{UP}}<0$.

\subsection{Evaluation Protocol}

We evaluate (i) AUROC/AUPRC for per-specification and any-violation labels,
(ii) association between disagreement and $|\hat\rho-\rho|$ (Pearson, descriptive),
and (iii) risk--coverage under disagreement-based abstention.

\section{Experimental Setup}
\label{sec:experiments}

\subsection{Environments}
For experiments we use MuJoCo~\cite{mujoco} Ant-v5, a 3D quadruped with eight actuated joints, and HalfCheetah-v5, a 2D planar locomotion system with six actuated joints.

\subsection{PPO Training}
We train PPO~\citep{ppo} policies with Stable-Baselines3~\citep{sb3} for 200,000 timesteps using five random seeds. The primary analysis uses three of these seeds, $\{0,42,123\}$, with 10 evaluation episodes per condition. We report descriptive means and standard deviations to summarize the observed seed-to-seed spread. We initialize each world-model ensemble member independently and train it on nominal trajectories. Appendix~\ref{app:sup-details} gives the full PPO and environment configurations.

\subsection{Distribution Shift and Prediction Horizons}
{\looseness=1

We evaluate external forces of 5 and 15 N, Gaussian observation
noise ($\sigma\in\{0.01,0.05,0.15\}$), mass scaling
($\{0.8,1.3,1.6\}$), friction scaling ($0.5\times$), and
compound shifts involving mass, noise, and force. All shifts
are absent from policy and world-model training.
We evaluate horizons $K\in\{1,5,10,20\}$, using windows within each
episode with stride $q=\max(1,K)$, so windows do not overlap in
transitions.
Early episode termination leads to variable sample counts.
Window-level correlations are descriptive because observations
within episodes are temporally dependent. Further configuration
and window-count details are given in Appendix~\ref{app:sup-primary}.

\subsection{Baselines}
The threshold and barrier-style monitors are retrospective references only
(App.~\ref{app:sup-legacy}); the barrier-style baseline is not a formal CBF.
Predictive experiments compare world-model predictions with observed-trajectory
scores, using disagreement baselines on identical rollouts: (B0) none, $S{=}\bar\rho$;
(B1) $U_o$; (B2) $U_z$; (B3) $U_\rho$. Conformal calibration is future work.
Error is reported as mean $L_2$ (within-environment), normalized RMSE,
safety-variable error, and $|\rho-\hat\rho|$.

\section{Results}
\label{sec:results}
We report predictive bias, failure prediction (robustness vs.\ disagreement), and horizon/shift sensitivity. The primary multi-seed failure-prediction analysis uses the window-level any-violation target; per-specification analyses diagnose the sources of these aggregate failures. Appendix~\ref{app:sup-legacy} reports the legacy single-seed analyses.

\subsection{Failure Prediction: Disagreement vs. Predicted Robustness}
\label{sec:extension}

The primary target is the window-level any-violation label,
\begin{equation}
Y_{i,t}^{(K)}
=\mathbf{1}[\rho_i(\tau_{t:t+K},\phi_i^K)<0],
\qquad
Y_t^{(K)}=\mathbf{1}[\exists i:Y_{i,t}^{(K)}=1].
\end{equation}
where $i$ indexes the evaluated STL specifications, $\rho_i$ is the
quantitative robustness of specification $i$ on the observed segment,
$\tau_{t:t+K}$ is that trajectory segment, $\phi_i^K$ is the specification
truncated to horizon $K$, and $\mathbf{1}[\cdot]$ is the indicator function. Thus,
$Y_t^{(K)}=1$ when at least one evaluated specification has negative
robustness over the window.
This target differs from the composite label
$Y_{\mathrm{comp},t}^{(K)}=\mathbf{1}[S_t^O(K)<0]$:
for example, the seed-0 severe-noise slice has 35.6\%
composite-label prevalence but 95.6\% any-violation prevalence.
At $K<50$, recovery contributes through its truncated
per-specification label but is excluded from the predictive
composite, as described in Sec.~\ref{sec:stl}.
Thus, AUROC for $-P$ measures how well predicted STL margins
rank violations of the corresponding observed STL specifications;
it is not an independent assessment of general physical safety.
We use $-P$ as a continuous failure-ranking score, whereas
$Y_t^{(K)}$ is a binary target.

We also evaluate observation- and latent-space disagreement, per-specification metrics, AUROC/AUPRC, risk--coverage, ensemble size, additional shifts, and a noise-model ablation. These experiments follow the protocol in Sec.~\ref{sec:experiments} (3 PPO seeds, 10 episodes per condition, $K{\in}\{10,20\}$, and $M{=}3$ with $M{=}5$/$M{=}10$ ablations). The replication package~\cite{replication} contains the per-window results and trained ensembles.

For HalfCheetah, at $K{<}50$ recovery is excluded from the predictive composite and the contact predicate is constant, contributing no within-environment discrimination (Sec.~\ref{sec:stl}). Per-specification diagnostics show observed failures concentrating in height and recovery; the latter contributes to the any-violation target through its truncated robustness, although it is excluded from the predictive composite.

\begin{table}[htbp]
\centering
\caption{Failure prediction at $K{=}10$ for the HalfCheetah any-violation target (mean$\pm$std of per-seed metrics over 3 PPO seeds, 10 episodes per seed). Viol\% is multi-seed any-violation prevalence, not comparable to seed-0 per-spec rates in Appendix~\ref{app:sup-primary}. Ant-v5 is not reported in this multi-seed table because its any-violation target is single-class ($79$--$100\%$ positive) in 21 of 24 evaluated seed-condition slices (8 conditions $\times$ 3 seeds); the three slices admitting both classes give AUROC($-P$) $0.98$--$1.00$ (Tab.~\ref{tab:ant_any}). Its composite target has no positive violations at all. Predicted robustness ($-P$) substantially outperforms the evaluated disagreement measures.}
\label{tab:auroc}
\begin{tabular}{lccccc}
\toprule
 & & \multicolumn{2}{c}{AUROC} & \multicolumn{2}{c}{AUPRC} \\
\cmidrule(lr){3-4}\cmidrule(lr){5-6}
Condition & Viol\% & $U_\rho$ & \textbf{$-P$} & $U_\rho$ & \textbf{$-P$} \\
\midrule
baseline & $56{\pm}25$ & $0.50{\pm}0.08$ & $\mathbf{0.92{\pm}0.08}$ & $0.59{\pm}0.26$ & $\mathbf{0.97{\pm}0.02}$ \\
noise 0.05 & $57{\pm}29$ & $0.50{\pm}0.07$ & $\mathbf{0.81{\pm}0.25}$ & $0.64{\pm}0.24$ & $\mathbf{0.95{\pm}0.05}$ \\
noise 0.15 & $56{\pm}29$ & $0.56{\pm}0.18$ & $\mathbf{0.98{\pm}0.00}$ & $0.61{\pm}0.28$ & $\mathbf{0.97{\pm}0.03}$ \\
mass 0.8x & $55{\pm}24$ & $0.47{\pm}0.09$ & $\mathbf{0.96{\pm}0.05}$ & $0.63{\pm}0.19$ & $\mathbf{0.98{\pm}0.02}$ \\
mass 1.6x & $63{\pm}28$ & $0.47{\pm}0.03$ & $\mathbf{0.94{\pm}0.01}$ & $0.47{\pm}0.04$ & $\mathbf{0.92{\pm}0.03}$ \\
friction 0.5x & $56{\pm}29$ & $0.58{\pm}0.13$ & $\mathbf{0.88{\pm}0.14}$ & $0.61{\pm}0.28$ & $\mathbf{0.96{\pm}0.04}$ \\
compound & $64{\pm}26$ & $0.45{\pm}0.02$ & $\mathbf{0.97{\pm}0.01}$ & $0.51{\pm}0.06$ & $\mathbf{0.97{\pm}0.00}$ \\
force 5N & $52{\pm}24$ & $0.51{\pm}0.07$ & $\mathbf{0.96{\pm}0.04}$ & $0.58{\pm}0.23$ & $\mathbf{0.97{\pm}0.02}$ \\
force 15N & $49{\pm}27$ & $0.52{\pm}0.07$ & $\mathbf{0.95{\pm}0.03}$ & $0.54{\pm}0.26$ & $\mathbf{0.93{\pm}0.06}$ \\
\bottomrule
\end{tabular}
\end{table}

The alternative observation- and latent-space disagreement
measures also show weak discrimination (AUROC 0.44--0.54
in the reported $K=10$ conditions). At $K=20$ under severe
noise, $U_\rho$, $U_o$, and $U_z$ achieve AUROC
$0.65\pm0.16$, $0.47\pm0.18$, and $0.46\pm0.07$,
respectively, compared with $0.99\pm0.02$ for $-P$. In the single-seed
(seed-0) composite-label horizon analysis, longer horizons expose more
violations without consistently strengthening disagreement as a failure-ranking
signal (Fig.~\ref{fig:horizon_sensitivity} and
Tab.~\ref{tab:horizon_auroc} in Appendix~\ref{app:sup-primary}); the multi-seed
any-violation result at $K{=}20$ is given above.



\pgfplotsset{compat=1.18}

\begin{figure*}[t]
\centering

\begin{tikzpicture}
\begin{axis}[
  width=0.55\textwidth,
  height=5.4cm,
  ybar,
  bar width=7pt,
  ymin=0.0, ymax=1.15,
  ylabel={AUROC (any-violation)},
  xlabel={Condition (HalfCheetah, $K{=}10$)},
  xtick={1,2,3,4},
  xticklabels={baseline, noise 0.15, mass 1.6x, force 15N},
  x tick label style={font=\scriptsize, rotate=15, anchor=east},
  y tick label style={font=\scriptsize},
  grid=major,
  grid style={gray!30},
  legend style={
    at={(0.02,0.98)},
    anchor=north west,
    font=\scriptsize,
    draw=none,
    fill=white,
    fill opacity=0.85
  },
  title={\textbf{Disagreement $\approx$ chance, $-P$ ranks failures}},
  title style={font=\small},
]
\addplot[gray!70, dashed, thick, forget plot]
  coordinates {(0.3,0.5) (4.7,0.5)};

\addplot[
  fill=red!55,
  error bars/.cd, y dir=both, y explicit,
  error bar style={black, thick},
] coordinates {
  (1,0.50) +- (0,0.08)
  (2,0.56) +- (0,0.18)
  (3,0.47) +- (0,0.03)
  (4,0.52) +- (0,0.07)
};
\addlegendentry{$U_\rho$}

\addplot[
  fill=blue!55,
  error bars/.cd, y dir=both, y explicit,
  error bar style={black, thick},
] coordinates {
  (1,0.92) +- (0,0.08)
  (2,0.98) +- (0,0.00)
  (3,0.94) +- (0,0.01)
  (4,0.95) +- (0,0.03)
};
\addlegendentry{$-P$}

\node[font=\scriptsize, gray!80, anchor=south] at (axis cs:2.5,0.53) {chance};
\end{axis}
\end{tikzpicture}
\hfill
\begin{tikzpicture}
\begin{axis}[
  width=0.36\textwidth,
  height=5.4cm,
  ybar,
  bar width=12pt,
  ymin=0.0, ymax=1.15,
  ylabel={AUROC (any-violation)},
  xlabel={Score (HC noise 0.15, $K{=}20$)},
  xtick={1,2,3,4},
  xticklabels={$U_\rho$, $U_o$, $U_z$, $-P$},
  x tick label style={font=\scriptsize},
  y tick label style={font=\scriptsize},
  grid=major,
  grid style={gray!30},
  title={\textbf{Pattern holds at $K{=}20$}},
  title style={font=\small},
]
\addplot[gray!70, dashed, thick, forget plot]
  coordinates {(0.3,0.5) (4.7,0.5)};

\addplot[
  fill=red!55!gray!10,
  error bars/.cd, y dir=both, y explicit,
  error bar style={black, thick},
] coordinates {
  (1,0.65) +- (0,0.16)
  (2,0.47) +- (0,0.18)
  (3,0.46) +- (0,0.07)
  (4,0.99) +- (0,0.02)
};
\end{axis}
\end{tikzpicture}

\caption{
HalfCheetah any-violation failure prediction (descriptive mean$\pm$std over 3 seeds).
\textbf{Left ($K{=}10$, Tab.~\ref{tab:auroc}):} $U_\rho$ stays near chance (0.47--0.56) while $-P$ ranks failures strongly (0.92--0.98); $U_o$/$U_z$ are likewise near chance (0.44--0.54).
\textbf{Right ($K{=}20$, severe noise, 64.2$\pm$24.4\% prevalence):} $U_\rho$/$U_o$/$U_z$ give $0.65$/$0.47$/$0.46$ versus $\mathbf{0.99}$ for $-P$. Dashed line marks chance.
}
\label{fig:failure_prediction}
\end{figure*}
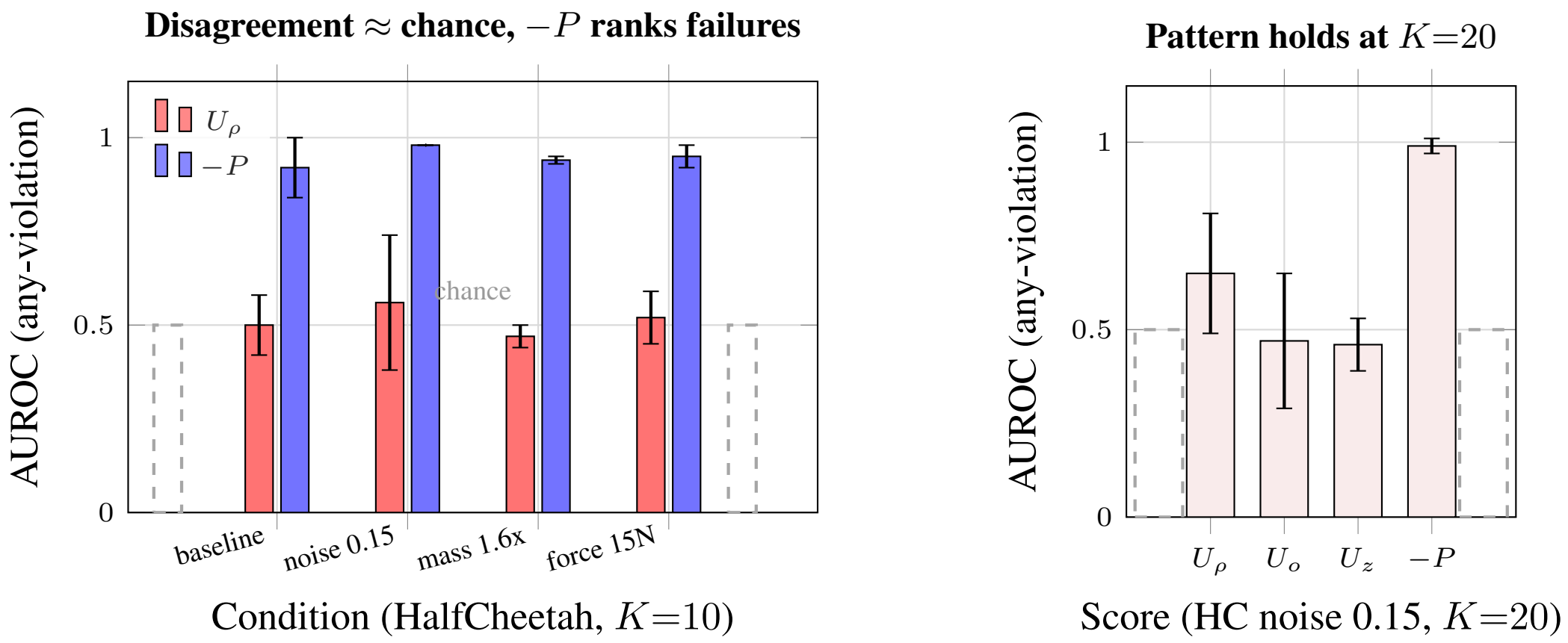


\textbf{Predicted robustness is more informative than disagreement.} A low predicted robustness margin $\bar\rho$ already provides useful safety information, and disagreement does not reliably add to it. The evaluated observation- and latent-space disagreement measures also perform inconsistently, so the lack of reliable failure discrimination is not specific to STL-space disagreement.

\textbf{Current-state baseline and early warning.}
Instantaneous robustness $-O_t$ (no prediction) reaches AUROC $0.79$--$0.97$,
within $0.03$ of $-P$ ($0.81$--$0.98$; Tab.~\ref{tab:auroc}). Among windows whose
initial state satisfies all pointwise predicates ($60$--$70\%$ of windows), future
violations still occur in $30$--$45\%$, yet $-O_t$ stays within $0.02$ AUROC of
$-P$ there, while $U_\rho$ is weaker and less consistent
($0.30$--$0.67$; Tab.~\ref{tab:current_state}, App.~\ref{app:sup-primary}).
Prediction thus adds little ranking information beyond the initial state in
these settings. Ant-v5 provides limited early-warning evidence because few
windows in its four originally evaluated conditions are safe at the start.

\textbf{Reliability, ensemble size, and noise-model checks.} Risk--coverage curves remain essentially flat under severe noise, with $U_\rho$-based abstention failing to reduce retained violation risk; increasing the ensemble size (ablations over $M{\in}\{2,3,5,10\}$) largely stabilizes the magnitude of $U_\rho$ without consistently improving failure discrimination; and matching the deployment noise process in rollout resolves neither the predictive bias nor disagreement-based failure detection (full numbers in Appendix~\ref{app:sup-primary}).

\textbf{Disagreement penalization does not improve failure ranking.}
On the composite-label target, increasing $\lambda$ reduces
AUROC from $0.88$ to $0.84$ under severe noise while increasing
the flag rate. On the any-violation target, penalization produces
only small AUROC changes ($0.983\pm0.003$ to
$0.989\pm0.004$ at $K=10$), which we do not interpret as a
robust improvement. Risk--coverage abstention also produces
little reduction in retained violation rate (55.9\% to 53.4\%
at 70\% coverage).
The flag rule $S_{\mathrm{UP}}{<}0$ uses the composite boundary, not the
any-violation label, so its recall is at most about $9\%$ ($5.2\%$ flag rate vs.\
$55.9\%$ prevalence); we do not present it as a calibrated operating point
(App.~\ref{app:sup-primary}).
Thus, neither penalization nor
disagreement-based abstention consistently improves failure
detection; complete sweeps are reported in Appendix~\ref{app:sup-primary}.

\section{Discussion}
\label{sec:discussion}
Retrospective STL evaluates robustness on observed trajectories, predictive STL estimates near-future robustness under the policy's closed-loop behavior, and ensemble STL disagreement measures the variation among predicted robustness values.

\textbf{Predictive STL exhibits environment-dependent bias.} Predictive scores are mildly pessimistic for Ant-v5 ($P$ 0.04 below $O$ at $K{=}10$) but systematically optimistic for HalfCheetah-v5 ($P$ exceeds $O$ by 0.01--0.06; Tab.~\ref{tab:retrospective_predictive} in Appendix~\ref{app:sup-legacy}). Neither sensing-vs-rollout mismatch nor matched-noise rollout ablation fully explains the bias (noise 0.15 bias $+0.12{\to}+0.13$), suggesting additional model-mismatch sources (Fig.~\ref{fig:predictive_bias} in Appendix~\ref{app:sup-legacy}).

\textbf{STL disagreement does not reliably predict either prediction error or future violations.} Across the three evaluated PPO seeds, single-seed correlations with prediction error flip sign at small $n$; the seed-averaged correlation is negative or near zero (HC $r_{U_\rho E}{=}-0.19{\pm}0.16$ at baseline and $-0.18{\pm}0.07$ under noise 0.15). On the HalfCheetah any-violation target, despite a substantial violation prevalence ($64.2{\pm}24.4\%$ at $K{=}20$ under noise), $U_\rho$ AUROC is $0.45$--$0.65$ across conditions, with AUPRC near prevalence. An above-chance exception appears in the Ant per-specification disagreement analysis (Appendix~\ref{app:sup-primary}), where $U_\rho$ reaches AUROC $0.73$--$0.83$ in two small-sample conditions (baseline and mass 1.6x). These small-sample results are descriptive and do not constitute a stable counterexample; they qualify the headline claim: disagreement is unreliable as a general safety signal, but not uniformly uninformative in every slice.}

\textbf{Predicted robustness outperforms disagreement, and penalization does not improve ranking.} On the HalfCheetah any-violation target at $K{=}10$, $-P$ substantially outperforms the evaluated disagreement measures (Tab.~\ref{tab:auroc}); disagreement penalization yields no robust improvement on either evaluation target (Sec.~\ref{sec:extension}). The relevant question is whether disagreement provides additional information about when the predicted robustness margin is unreliable.

We train world models only on nominal trajectories and do not adapt them at test time; the results therefore characterize disagreement under shift in this fixed-model setting. Here, $U_\rho$ is a condition- and horizon-dependent measure that does not provide a generally reliable indication of predictive reliability or safety. This finding is scoped to lightweight feedforward ensembles, whose independently initialized members may still exhibit correlated errors or shared bias; whether recurrent or stochastic-latent (Dreamer-style) ensembles yield more informative disagreement remains for future work. The ensemble quantity measures STL robustness disagreement, not aleatoric uncertainty; estimating the latter requires explicit conditional-variance predictions.

\section{Limitations}
\label{sec:limitations}
The current study has several limitations. First, evaluation
is restricted to two MuJoCo environments and manually defined
STL specifications. The specifications are more naturally
aligned with Ant-v5, while quantitative failure-prediction
analysis is centered on HalfCheetah-v5. Sensitivity to
alternative specification sets, aggregation weights, and
windowing strides remains untested. Second, the world model is a lightweight feedforward predictor,
and ensemble disagreement is an empirical measure rather than
calibrated uncertainty or a formal safety guarantee. Evaluation
uses three PPO seeds, and the $M=3$ disagreement estimate may
remain sensitive to ensemble composition despite the reported
ensemble-size analysis. Because $-P$ evaluates predicted states against the same
STL predicates used to construct the observed-trajectory label, its advantage
over the unsigned spread $U_\rho$ is partly structural; we do not test whether
$U_\rho$ adds information conditional on $-P$. Finally, simulation does not capture real-world state-estimation,
actuation, contact-model, or latency effects. Predictive
monitoring also incurs additional computation because it
requires multiple model rollouts.

\section{Conclusion}
\label{sec:conclusion}

This study evaluated predictive STL monitoring of frozen
locomotion policies under distribution shift using
closed-loop world-model rollouts and ensemble disagreement.
Across the evaluated settings, disagreement did not reliably
predict model error or future STL violations, whereas
predicted robustness provided stronger failure ranking.
However, instantaneous current-state robustness already
captured much of this information, limiting the additional
ranking benefit of prediction. Penalizing robustness by
disagreement also failed to consistently improve failure ranking. These findings
caution against treating ensemble disagreement alone as a reliable indicator of
safety-relevant uncertainty in the evaluated settings.
The evaluation is limited to two simulated MuJoCo
environments and three PPO seeds
(Sec.~\ref{sec:limitations}). Future work will investigate
conformal calibration and extend predictive monitoring
to real-world robotic systems.

\bibliography{references}

@inproceedings{stl,
  title={Monitoring Temporal Properties of Continuous Signals},
  author={Maler, Oded and Nickovic, Dejan},
  booktitle={Formal Modeling and Analysis of Timed Systems (FORMATS)},
  year={2004}
}

@article{yamaguchi2024rtamt,
  title={{RTAMT} --- Runtime Robustness Monitors with Application to {CPS} and Robotics},
  author={Yamaguchi, Tomoya and Hoxha, Bardh and Ni\v{c}kovi\'{c}, Dejan},
  journal={International Journal on Software Tools for Technology Transfer},
  year={2023}
}

@inproceedings{donze2011robust,
  title={Robust Satisfaction of Temporal Logic over Real-Valued Signals},
  author={Donz{\'{e}}, Alexandre and Maler, Oded},
  booktitle={Formal Modeling and Analysis of Timed Systems (FORMATS)},
  series={Lecture Notes in Computer Science},
  volume={6246},
  pages={92--106},
  year={2010}
}

@article{meng2023stlnpc,
  title={Signal Temporal Logic Neural Predictive Control},
  author={Meng, Yue and Fan, Chuchu},
  journal={IEEE Robotics and Automation Letters},
  year={2023}
}

@article{cairoli2025conformal,
  title={Conformal Quantitative Predictive Monitoring of Stochastic Systems with Conditional Validity},
  author={Cairoli, Francesca and Kuipers, Tom and Bortolussi, Luca and Paoletti, Nicola},
  journal={Nonlinear Analysis: Hybrid Systems},
  year={2025}
}

@article{han2025mpmonitoring,
  title={Model Predictive Online Monitoring of Dynamical Systems for Nested Signal Temporal Logic Specifications},
  author={Han, Tao and Li, Shaoyuan and Yin, Xiang},
  journal={arXiv preprint arXiv:2505.19703},
  year={2025}
}

@article{predictive_safety,
  title={Conformal Prediction for {STL} Runtime Verification},
  author={Lindemann, Lars and Qin, Xu and Deshmukh, Jyotirmoy V. and Pappas, George J.},
  journal={arXiv preprint arXiv:2211.01539},
  year={2022}
}

@inproceedings{zhao2024conformalstl,
  title={Robust Conformal Prediction for {STL} Runtime Verification under Distribution Shift},
  author={Zhao, Yiqi and Hoxha, Bardh and Fainekos, Georgios and Deshmukh, Jyotirmoy V. and Lindemann, Lars},
  booktitle={ACM/IEEE International Conference on Cyber-Physical Systems (ICCPS)},
  year={2024}
}

@inproceedings{beirami2026conformal,
  title={Conformal Signal Temporal Logic for Robust Reinforcement Learning Control: A Case Study},
  author={Beirami, Hani and Islam, M M Manjurul},
  booktitle={ICASSP 2026: IEEE International Conference on Acoustics, Speech and Signal Processing},
  pages={5881--5885},
  year={2026},
  organization={IEEE}
}

@inproceedings{ensemble,
  title={Simple and Scalable Predictive Uncertainty Estimation using Deep Ensembles},
  author={Lakshminarayanan, Balaji and Pritzel, Alexander and Blundell, Charles},
  booktitle={Advances in Neural Information Processing Systems (NeurIPS)},
  year={2017}
}

@article{ppo,
  title={Proximal Policy Optimization Algorithms},
  author={Schulman, John and Wolski, Filip and Dhariwal, Prafulla and Radford, Alec and Klimov, Oleg},
  journal={arXiv preprint arXiv:1707.06347},
  year={2017}
}

@article{sb3,
  title={{Stable-Baselines3}: Reliable Reinforcement Learning Implementations},
  author={Raffin, Antonin and Hill, Ashley and Ernestus, Maximilian and Berber, Tristan and Gleave, Adam and Kanervisto, Anssi and Dormann, Nicolas},
  journal={Journal of Machine Learning Research},
  year={2021}
}

@article{dreamer,
  title={Mastering Diverse Domains through World Models},
  author={Hafner, Danijar and Lillicrap, Timothy and Ha, Jimmy and Norouzi, Mohammad},
  journal={arXiv preprint arXiv:2301.04104},
  year={2023}
}

@inproceedings{mujoco,
  title={{MuJoCo}: A Physics Engine for Model-Based Control},
  author={Todorov, Emanuel and Erez, Tom and Tassa, Yuval},
  booktitle={IEEE/RSJ International Conference on Intelligent Robots and Systems (IROS)},
  year={2012}
}

@inproceedings{cobbe2020quantifying,
  title={Quantifying Generalization in Reinforcement Learning},
  author={Cobbe, Karl and Hesse, Chris and Hilton, Jacob and Schulman, John},
  booktitle={International Conference on Machine Learning (ICML)},
  year={2020}
}

@misc{antv5,
  title={{Gymnasium} {MuJoCo} v5 Environments: {Ant}},
  author={{Farama Foundation}},
  year={2024},
  url={https://farama.org/Gymnasium-MuJoCo-v5_Environments},
  note={Accessed: 2026-09-11}
}

@misc{halfcheetahv5,
  title={Half Cheetah: {Gymnasium} {MuJoCo} Environment},
  author={{Farama Foundation}},
  year={2024},
  url={https://gymnasium.farama.org/environments/mujoco/half_cheetah/},
  note={Accessed: 2026-09-11}
}

@misc{replication,
	title        = {Replication Package},
	author       = {GitHub},
	year         = 2026,
	url          = {https://github.com/lakhanqurban/stl-monitor-rl},
	note         = {Accessed: 2026-11-01}
}

\clearpage
\appendix

\section{Supplementary Experimental Details}\label{app:sup-details}

\subsection{Safety-Variable Mapping}

The STL predicates are evaluated on a safety-variable representation
$x_t=q(o_t)$ derived from the environment observation. The function
$q(\cdot)$ extracts or computes the physical variables required by the STL
specifications from the raw observation vector. For Ant-v5, $q(\cdot)$
extracts body height from the position components, computes roll and pitch
angles from the torso quaternion, and extracts forward velocity from the
linear velocity components. For HalfCheetah-v5, $q(\cdot)$ extracts the
variables required by the applicable specifications; height and pitch come
from the observation, roll is set to zero because the environment is planar,
and the airborne count is fixed at 2.

The mapping $q(\cdot)$ is implemented using the MuJoCo observation
definitions for each environment and is applied identically to observed and
predicted observations. This ensures that retrospective and predictive
robustness are evaluated in the same safety-variable space, making the two
quantities directly comparable.

\subsection{Episode-Level Specifications}

The retrospective specifications used for full-episode evaluation are
defined over a 500-transition episode, represented by up to 501 state
observations under the inclusive interval notation, and capture the
safety requirements for locomotion:
\begin{align}
\phi_1&=G_{[0,500]}(|roll|<1.0),\\
\phi_2&=G_{[0,500]}(|pitch|<1.0),\\
\phi_3&=G_{[0,500]}(height>0.4),\\
\phi_4&=G_{[0,500]}(airborne\leq3),\\
\phi_5&=G_{[0,500]}(|v_{\mathrm{fwd}}|<4.0),\\
\phi_6&=G_{[0,500]}(height\leq0.45\rightarrow F_{[0,50]}(height>0.5)).
\end{align}
Specifications $\phi_1$ and $\phi_2$ constrain body orientation to prevent
excessive tilting. Specification $\phi_3$ ensures the body remains above a
minimum height to prevent ground contact. Specification $\phi_4$ limits the
number of airborne contacts to maintain stable locomotion. Specification
$\phi_5$ bounds forward velocity to prevent destabilizing speeds.
Specification $\phi_6$ encodes a recovery property: if the body drops below
a threshold, it must recover within 50 steps.

For online evaluation, the specifications are converted to past-time form.
Under truncated $K$-step evaluation with $K<50$, the recovery robustness
is computed using the available trace segment and the corresponding
past-time semantics. Such a value characterizes the observed prefix and
does not establish whether the complete 50-step recovery obligation has
been met. Accordingly, truncated recovery values are retained as
per-specification diagnostics, but should not be interpreted as
evaluations of the complete 50-step obligation. A complete evaluation
requires the trace to cover the entire response interval under the
implementation's discrete-time indexing. If the response interval starts
at the initial state and spans 50 transitions, this corresponds to
$K\geq50$; the exact boundary must be checked against the past-time
recurrence and the indexing used by the evaluator (Sec.~\ref{sec:stl}).
The $K$-step rollout contains $K$ transitions and $K+1$ state
observations when both its initial and final states are included.

These specifications are designed primarily around Ant-v5; cross-environment
interpretation should be treated cautiously. For HalfCheetah-v5, the
applicable specifications are pitch, height, contact, and recovery (roll is
omitted because the environment is planar and forward velocity is excluded
as non-safety-critical). Under the adopted HalfCheetah mapping, the
airborne count is fixed at 2, so the contact predicate
($airborne\leq3$) has constant positive robustness and provides no
within-environment discrimination. Nevertheless, retaining this term in
an arithmetic composite contributes a constant positive offset and can
therefore shift the composite's zero boundary. HalfCheetah composite
violation rates should consequently be interpreted with this constant
contribution in mind. A positive composite robustness score does not imply
that every individual specification is satisfied: positive robustness
contributions from some predicates can offset negative contributions from
others.

Composite weights are
$\mathbf{w}_{\mathrm{retro}}=[1.0,1.0,0.8,0.5,1.0,0.8]$ for the six
retrospective properties and
$\mathbf{w}=[1.0,1.0,0.8,0.5,1.0]$ for the five Ant-v5 predictive
properties. Raw robustness values have incompatible physical units (m, rad,
m/s), so composites are environment-specific monitoring indices, not
absolute safety quantities.

\subsection{PPO Configuration}

PPO policies are trained with Stable-Baselines3 for 200,000 timesteps using
seeds $\{0,42,123,456,789\}$: learning rate $3\times10^{-4}$, rollout
length 2048, batch size 64, 10 epochs, GAE $0.95$, discount $0.99$, clip
range $0.2$, with VecNormalize for observations and rewards. The primary
evaluation aggregates seeds $\{0,42,123\}$ (10 episodes per condition);
the remaining two seeds are not included in the primary aggregate.
Reported mean$\pm$std values are descriptive spread over the three seeds
(not standard errors; with three points, e.g.\ $0.81{\pm}0.25$ reports
range-like variability, not a sampling distribution).

\subsection{World-Model Training}

We train an ensemble of $M{=}3$ lightweight learned world models per
environment as the main configuration (a feedforward latent dynamics model
used as a predictive world model, not a full-scale Dreamer system), extended
to $M{=}5$ and $M{=}10$ with identically trained additional members
(Sec.~\ref{sec:extension}; Ant final evaluation MSE $0.031$--$0.040$,
HalfCheetah $0.0018$--$0.0032$). Each member is a feedforward network: an
encoder maps the 105-dimensional (Ant) or 17-dimensional (HalfCheetah)
observation to a 128-dimensional latent state, a dynamics model with two
hidden layers of 256 units (LayerNorm, SiLU) predicts the next latent state
given the current latent state and action, and a decoder reconstructs the
observation. A reward head is included as an auxiliary prediction task,
though reward is not used by the safety monitor.

Models are trained on up to 25{,}000 nominal transitions from 50 episodes of
up to 500 steps each collected from a trained PPO policy. The training loss
combines observation reconstruction and auxiliary reward prediction
~\cite{dreamer}:
\begin{equation}
\mathcal{L}_m=\frac{1}{N}\sum_{n=1}^{N}
\left[
\|g_\theta(\hat{z}_{n+1})-o_{n+1}\|_2^2
+\beta\cdot(r_n-\hat{r}_n)^2
\right],
\end{equation}
where $N$ is the number of training transitions, $n$ indexes a transition,
and $m$ indexes an ensemble member. The quantities $o_n$, $a_n$, and $r_n$
are the observed observation, action, and reward; $\hat{r}_n$ is the
predicted reward. The predicted latent state is
$\hat{z}_{n+1}=d_\theta(e_\theta(o_n),a_n)$, and
$g_\theta(\hat{z}_{n+1})$ is the decoded observation. The parameter
$\beta{=}0.1$ weights the reward prediction term. Training uses 30 epochs,
learning rate $10^{-3}$, and batch size 256, with ensemble members using
different random seeds. Training loss (MSE) and evaluation error ($L_2$
observation error across closed-loop rollouts) are reported under different
aggregation and are not directly comparable; final training losses are
0.017, 0.022, and 0.037 for Ant-v5 and 0.002, 0.002, and 0.004 for
HalfCheetah-v5.

\section{Supplementary Legacy Single-Seed Results}\label{app:sup-legacy}

The results in this section use a single-seed (seed-0) evaluation protocol
retained from the original analysis and are provided for descriptive
comparison; quantitative claims in the main text and in
Appendix~\ref{app:sup-primary} are based on the multi-seed protocol, and
numerical values from the two protocols are not interchangeable.

\subsection{Retrospective Baseline}

The retrospective evaluation reported an Ant-v5 composite STL score of
$+0.027$ nominally and $-0.202$ under severe observation noise, evaluated
over full 500-step episodes using all six specifications. The nominal
episode-level composite STL violation rate was about 40\%, compared with
about 6\% for the threshold and barrier-style monitors. This rate is
computed over full-episode trajectories and is not directly comparable to
the per-window violation rates reported below. These episode-level mean
robustness values are continuous averages and should not be interpreted as
episode-level violation rates; in particular, a negative mean composite
robustness does not imply that every episode violates the composite
specification.

\subsection{Predictive Monitoring}

Table~\ref{tab:summary} summarizes the key predictive monitoring metrics
across conditions at $K{=}10$, using closed-loop policy-conditioned
world-model rollouts. The predictive composite excludes the recovery
property ($\phi_6$) because it requires a 50-step temporal interval, which
is longer than the evaluated prediction horizon. Ant-v5 uses five
predictive properties. HalfCheetah-v5 primary evaluation uses the
applicable predictive properties available at each horizon; the constant
contact predicate is retained in the composite sum but excluded from
failure-prediction targets (a constant predicate cannot discriminate), so
the HalfCheetah predictive composite comprises pitch, height, and contact.
Per-specification violation rates are primary; composites are secondary.

\begin{table*}[t]
\centering
\caption{Legacy single-seed predictive monitoring summary at $K{=}10$
(closed-loop, environment-dependent property composite). $O$: retrospective
reference score evaluated on the actually observed future trajectory.
$P$: predictive score evaluated on the world-model trajectory.
$\delta_\rho$: mean $|\Delta_\rho|$. $U_\rho$: mean STL robustness
disagreement. $r_{UE}$: Pearson correlation between STL robustness
disagreement and prediction error ($n$ denotes monitoring windows; small
$n$ values result from early episode termination in shifted conditions).
Viol\%: future violation rate.}
\label{tab:summary}
\begin{tabular}{lccccccc}
\toprule
& \multicolumn{3}{c}{Composite STL} & & & & \\
\cmidrule(lr){2-4}
Condition & $O$ & $P$ & $\delta_\rho$ & $U_\rho$ & $r_{UE}$ ($n$) & $r_{UV}$ & Viol\% \\
\midrule
\multicolumn{8}{l}{\textit{Ant-v5}} \\
baseline     & $+0.75$ & $+0.71$ & $0.04$ & $0.049$ & $-0.75$ (3) & --- & 0.0 \\
force 5N     & $+0.75$ & $+0.73$ & $0.02$ & $0.056$ & $+0.49$ (3) & --- & 0.0 \\
force 15N    & $+0.74$ & $+0.72$ & $0.07$ & $0.043$ & $-0.36$ (19) & --- & 0.0 \\
noise 0.05   & $+0.66$ & $+0.70$ & $0.05$ & $0.035$ & $-0.95$ (5) & --- & 0.0 \\
noise 0.15   & $+0.60$ & $+0.70$ & $0.10$ & $0.037$ & $+0.06$ (9) & --- & 0.0 \\
mass 0.8x    & $+0.77$ & $+0.74$ & $0.04$ & $0.052$ & $-0.72$ (3) & --- & 0.0 \\
mass 1.6x    & $+0.66$ & $+0.69$ & $0.05$ & $0.038$ & $-0.47$ (10) & --- & 0.0 \\
\midrule
\multicolumn{8}{l}{\textit{HalfCheetah-v5}} \\
baseline     & $+0.11$ & $+0.14$ & $0.04$ & $0.015$ & $-0.28$ (27) & $-0.20$ & 7.4 \\
force 5N     & $+0.12$ & $+0.17$ & $0.06$ & $0.017$ & $-0.15$ (27) & $-0.26$ & 7.4 \\
force 15N    & $+0.08$ & $+0.13$ & $0.06$ & $0.009$ & $-0.12$ (27) & $-0.08$ & 7.4 \\
noise 0.05   & $+0.18$ & $+0.20$ & $0.07$ & $0.015$ & $-0.16$ (27) & $-0.14$ & 11.1 \\
noise 0.15   & $+0.03$ & $+0.15$ & $0.12$ & $0.010$ & $-0.09$ (27) & $+0.27$ & 37.0 \\
mass 0.8x    & $+0.12$ & $+0.18$ & $0.06$ & $0.021$ & $+0.00$ (27) & $+0.08$ & 3.7 \\
mass 1.6x    & $+0.05$ & $+0.15$ & $0.10$ & $0.012$ & $+0.18$ (27) & $-0.23$ & 11.1 \\
\bottomrule
\end{tabular}
\end{table*}

\subsection{World-Model Prediction Accuracy}

Table~\ref{tab:pred_error} reports the ensemble mean observation prediction
errors averaged across all conditions for each horizon, using closed-loop
rollouts. Training loss (MSE) and evaluation error ($L_2$ observation error
across the $K$-step closed-loop rollout) are reported under different
aggregation and are not directly comparable.

\begin{table}[t]
\centering
\caption{Mean observation prediction error ($L_2$) by prediction horizon,
closed-loop rollouts, averaged over all conditions. $^*$Ant-v5 $K{=}20$
is reported only for the subset of conditions with sufficient valid
monitoring windows.}
\label{tab:pred_error}
\begin{tabular}{lrrrr}
\toprule
Environment & $K{=}1$ & $K{=}5$ & $K{=}10$ & $K{=}20$ \\
\midrule
Ant-v5         & 0.30 & 0.31 & 0.32 & 0.34$^*$ \\
HalfCheetah-v5 & 2.52 & 2.49 & 2.48 & 2.50 \\
\bottomrule
\end{tabular}
\end{table}

Ant-v5 prediction error grows gradually from 0.30 ($K{=}1$) to 0.32
($K{=}10$), consistent with modest error accumulation, though $K{=}20$ is
reported only for the subset of conditions with sufficient valid monitoring
windows (force 15N, noise 0.15, mass 1.6x). HalfCheetah-v5 exhibits $L_2$
errors roughly $8\times$ larger than Ant-v5, while the error stays
relatively stable across horizons. Absolute $L_2$ errors should be
interpreted within each environment rather than compared across
environments, due to differences in state-space scale.

\subsection{Retrospective versus Predictive STL}

Table~\ref{tab:retrospective_predictive} compares retrospective and
predictive composite STL scores at each horizon for the nominal condition,
using the environment-dependent predictive composite. The results reveal
environment-dependent differences between retrospective and predictive
scores. For Ant-v5 at $K{=}1$, retrospective and predictive scores are
nearly identical, as expected for a single-step prediction, while at
$K{=}10$, $P$ is 0.04 below $O$, indicating a mildly pessimistic
predictive robustness estimate. For the nominal HalfCheetah-v5 condition,
the predictive score exceeds the retrospective score by 0.01--0.06 across
the evaluated horizons (Tab.~\ref{tab:retrospective_predictive}), indicating
an optimistic difference in this configuration.

\begin{table}[t]
\centering
\caption{Retrospective ($O$) versus predictive ($P$) composite STL score
at each horizon, nominal condition, closed-loop. Ant-v5 uses five
predictive properties; HalfCheetah-v5 uses three (pitch, height, contact;
recovery excluded at $K{<}50$).}
\label{tab:retrospective_predictive}
\begin{tabular}{lccccc}
\toprule
& \multicolumn{4}{c}{Horizon $K$} \\
\cmidrule(lr){2-5}
Environment & Metric & $K{=}1$ & $K{=}5$ & $K{=}10$ & $K{=}20$ \\
\midrule
\multirow{2}{*}{Ant-v5}
& $O$ & $+0.82$ & $+0.77$ & $+0.75$ & --- \\
& $P$ & $+0.82$ & $+0.78$ & $+0.71$ & --- \\
\midrule
\multirow{2}{*}{HalfCheetah-v5}
& $O$ & $+0.17$ & $+0.13$ & $+0.11$ & $+0.10$ \\
& $P$ & $+0.18$ & $+0.16$ & $+0.14$ & $+0.16$ \\
\bottomrule
\end{tabular}
\end{table}

\pgfplotsset{compat=1.18}

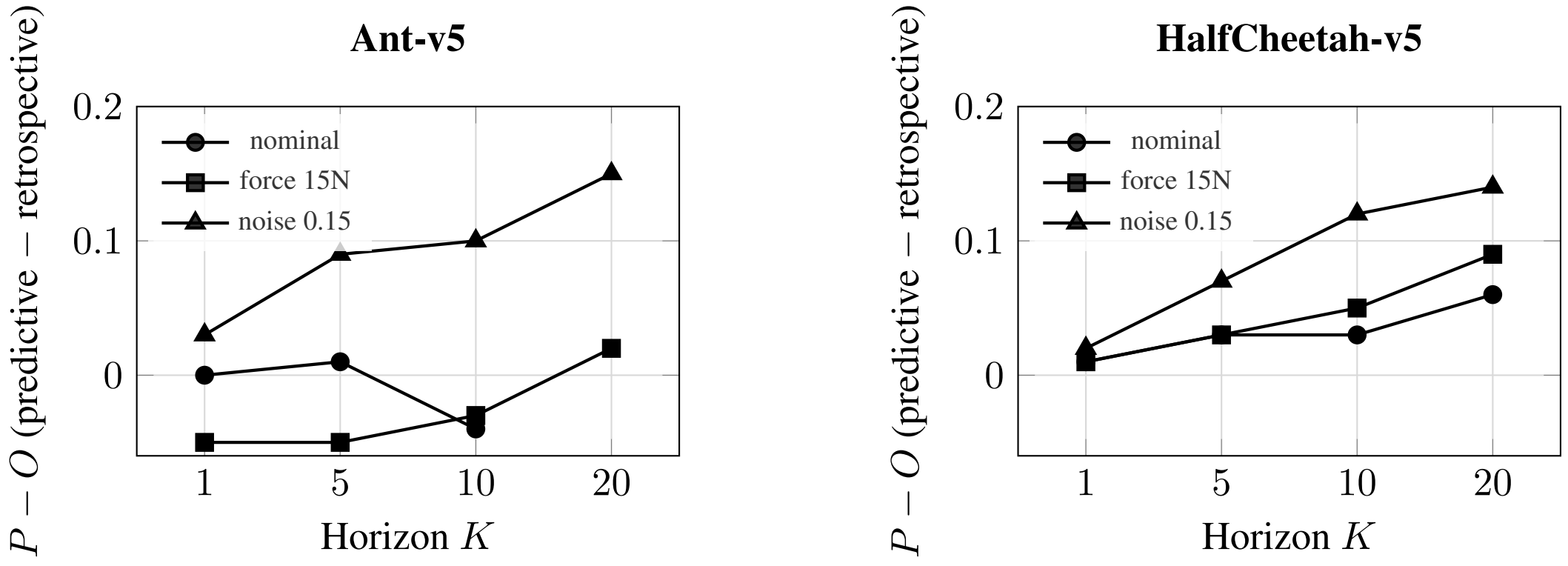
\begin{figure*}[t]
\centering

\begin{tikzpicture}
\begin{axis}[
  width=0.46\textwidth,
  height=4.7cm,
  xlabel={Horizon $K$},
  ylabel={$P-O$ (predictive $-$ retrospective)},
  xmin=0.5, xmax=4.5,
  xtick={1,2,3,4},
  xticklabels={$1$,$5$,$10$,$20$},
  ymin=-0.06, ymax=0.20,
  grid=major,
  grid style={gray!30},
  legend style={
    at={(0.02,0.98)},
    anchor=north west,
    font=\scriptsize,
    draw=none,
    fill=white,
    fill opacity=0.8
  },
  title={\textbf{Ant-v5}},
  title style={at={(0.5,1.05)}, anchor=south},
]

\addplot[
  mark=*,
  mark size=2pt,
  thick
] coordinates {
  (1, 0.00)
  (2, 0.01)
  (3, -0.04)
};
\addlegendentry{nominal}

\addplot[
  mark=square*,
  mark size=2pt,
  thick
] coordinates {
  (1, -0.05)
  (2, -0.05)
  (3, -0.03)
  (4, 0.02)
};
\addlegendentry{force 15N}

\addplot[
  mark=triangle*,
  mark size=2.5pt,
  thick
] coordinates {
  (1, 0.03)
  (2, 0.09)
  (3, 0.10)
  (4, 0.15)
};
\addlegendentry{noise 0.15}

\end{axis}
\end{tikzpicture}
\hfill
\begin{tikzpicture}
\begin{axis}[
  width=0.46\textwidth,
  height=4.7cm,
  xlabel={Horizon $K$},
  ylabel={$P-O$ (predictive $-$ retrospective)},
  xmin=0.5, xmax=4.5,
  xtick={1,2,3,4},
  xticklabels={$1$,$5$,$10$,$20$},
  ymin=-0.06, ymax=0.20,
  grid=major,
  grid style={gray!30},
  legend style={
    at={(0.02,0.98)},
    anchor=north west,
    font=\scriptsize,
    draw=none,
    fill=white,
    fill opacity=0.8
  },
  title={\textbf{HalfCheetah-v5}},
  title style={at={(0.5,1.05)}, anchor=south},
]

\addplot[
  mark=*,
  mark size=2pt,
  thick
] coordinates {
  (1, 0.01)
  (2, 0.03)
  (3, 0.03)
  (4, 0.06)
};
\addlegendentry{nominal}

\addplot[
  mark=square*,
  mark size=2pt,
  thick
] coordinates {
  (1, 0.01)
  (2, 0.03)
  (3, 0.05)
  (4, 0.09)
};
\addlegendentry{force 15N}

\addplot[
  mark=triangle*,
  mark size=2.5pt,
  thick
] coordinates {
  (1, 0.02)
  (2, 0.07)
  (3, 0.12)
  (4, 0.14)
};
\addlegendentry{noise 0.15}

\end{axis}
\end{tikzpicture}

\caption{
Predictive bias $(P-O)$ versus prediction horizon for representative
conditions. Ant-v5 exhibits mixed bias directions: predictive robustness
is mildly pessimistic under the 15N force shift but increasingly optimistic
under severe observation noise. HalfCheetah-v5 is consistently optimistic,
with the predictive--retrospective gap increasing with horizon and becoming
largest under observation noise.
}
\label{fig:predictive_bias}
\end{figure*}

Under distribution shift, the gap between retrospective and predictive
scores can increase. For HalfCheetah-v5 under noise $\sigma{=}0.15$, the
retrospective score at $K{=}20$ drops to $+0.01$ while the predictive score
remains at $+0.15$ (bias $P{-}O{\approx}+0.14$,
Fig.~\ref{fig:predictive_bias}; Tab.~\ref{tab:lcb} reports the same
$+0.15$ predictive score). This illustrates substantial optimistic
difference between the predictive and retrospective scores under severe
observation noise. The world model receives the noisy observation at the
start of the rollout, but subsequent predicted observations are generated
by the learned dynamics model rather than by the deployment-time
observation-noise process. The closed-loop rollout may therefore fail to
reproduce the persistent sensing-induced action perturbations present in
the actual trajectory.

\subsection{STL Disagreement versus Prediction Error}

Table~\ref{tab:uncertainty_error} reports the Pearson correlation between
STL robustness disagreement $U_\rho$ and prediction error $E_t$ by
condition and horizon. Correlations are computed over monitoring windows;
these windows are temporally dependent within episodes and should be
interpreted descriptively rather than as tests of statistical significance. In particular, several Ant-v5 correlations are based on only
3--5 windows; their magnitude and sign are highly sensitive to
individual observations and should be treated as illustrative
rather than representative of a condition-level relationship.

\begin{table}[t]
\centering
\caption{Pearson correlation ($r$) between STL robustness disagreement
and observation prediction error, closed-loop. All values are descriptive;
correlations with small sample sizes ($n<10$) should be interpreted with
particular caution.}
\label{tab:uncertainty_error}
\begin{tabular}{lcccc}
\toprule
& \multicolumn{4}{c}{Horizon $K$} \\
\cmidrule(lr){2-5}
Condition & $K{=}1$ & $K{=}5$ & $K{=}10$ & $K{=}20$ \\
\midrule
\multicolumn{5}{l}{\textit{Ant-v5}} \\
baseline     & $+0.10$ & $-0.58$ & $-0.75$ & --- \\
force 5N     & $+0.13$ & $-0.85$ & $+0.49$ & --- \\
force 15N    & $-0.12$ & $-0.29$ & $-0.36$ & $-0.39$ \\
noise 0.05   & $+0.12$ & $-0.05$ & $-0.95$ & --- \\
noise 0.15   & $-0.13$ & $-0.38$ & $+0.06$ & $+0.39$ \\
mass 0.8x    & $-0.09$ & $-0.56$ & $-0.72$ & --- \\
mass 1.6x    & $-0.07$ & $-0.28$ & $-0.47$ & $-0.98$ \\
\midrule
\multicolumn{5}{l}{\textit{HalfCheetah-v5}} \\
baseline     & $-0.14$ & $-0.05$ & $-0.28$ & $-0.55$ \\
force 5N     & $-0.14$ & $+0.01$ & $-0.15$ & $-0.59$ \\
force 15N    & $-0.04$ & $+0.06$ & $-0.12$ & $-0.38$ \\
noise 0.05   & $-0.11$ & $+0.01$ & $-0.16$ & $-0.07$ \\
noise 0.15   & $-0.05$ & $+0.14$ & $-0.09$ & $+0.65$ \\
mass 0.8x    & $-0.08$ & $+0.11$ & $+0.00$ & $-0.02$ \\
mass 1.6x    & $+0.04$ & $+0.26$ & $+0.18$ & $-0.04$ \\
\bottomrule
\end{tabular}
\end{table}

Several Ant-v5 conditions exhibit negative associations between STL
robustness disagreement and prediction error at $K{=}5$ and $K{=}10$.
Larger STL-robustness disagreement is associated with lower mean prediction
error in these conditions. One possible interpretation is that ensemble
disagreement reflects variation in local dynamical complexity rather than
prediction error itself. However, the present experiments do not directly
test this explanation. Within-episode temporal dependence limits the
interpretation of these correlations, and the observed sign changes should
not be treated as evidence of horizon-dependent effects.

For conditions with more monitoring windows, including force 15N
($n{=}19$), mass 1.6x ($n{=}10$), and HalfCheetah ($n{=}27$), the observed
correlations show heterogeneity across shift type, environment, and horizon
(e.g.\ $r{=}-0.55$ nominal, $r{=}-0.59$ under 5N force, and
$r{=}+0.65$ under observation noise $\sigma{=}0.15$ at $K{=}20$). The two
quantities capture different aspects of predictive mismatch: $E_t$
measures discrepancy in the full observation representation, whereas
$U_\rho$ measures disagreement after mapping predictions into the
safety-variable representation and STL robustness space.

\subsection{Disagreement Measures and Evaluation Details}
\label{app:sup-disagreement}

\paragraph{Prediction error.}
For each ensemble member $m$, we calculate the mean observation-space
prediction error over the $K$-step rollout:
\begin{equation}
E_t^{(m)} =
\frac{1}{K}\sum_{k=1}^{K}
\|\hat{o}^{(m)}_{t+k}-o_{t+k}\|_2,
\end{equation}
where $m$ indexes an ensemble member, $\hat{o}^{(m)}_{t+k}$ and $o_{t+k}$
are its predicted and the observed future observations, respectively, and
$K$ is the number of predicted steps.
The per-member errors are averaged across the ensemble:
\begin{equation}
E_t=\frac{1}{M}\sum_{m=1}^{M}E_t^{(m)}.
\end{equation}
where $M$ is the number of ensemble members and $E_t$ is their mean
observation-space prediction error.
This quantity serves as an empirical reference for prediction
accuracy. It is not a safety metric because it measures discrepancy
in observation space rather than in the safety-variable space
specified by STL.

\paragraph{Alternative disagreement representations.}
In addition to STL robustness disagreement $U_\rho$, we compute
observation-space disagreement $U_o$ and latent-space disagreement
$U_z$ on the same predicted rollouts:
\begin{equation}
U_o(t)=
\frac{1}{K}\sum_{k=1}^{K}
\operatorname{mean}_j
\operatorname{Std}_m
\left(\hat{o}^{(m)}_{t+k,j}\right),
\end{equation}
\begin{equation}
U_z(t)=
\frac{1}{K}\sum_{k=1}^{K}
\operatorname{mean}_j
\operatorname{Std}_m
\left(\hat{z}^{(m)}_{t+k,j}\right),
\end{equation}
where $m$ indexes ensemble members, $j$ indexes observation or latent
components, $K$ is the rollout horizon, and $\operatorname{Std}_m$ denotes
the standard deviation across ensemble members. The component-wise standard deviations are
averaged over representation dimensions and rollout time steps.
These measures allow us to examine whether the observed
failure-prediction behavior is specific to the STL representation
or is also present in observation and latent spaces.

\paragraph{Disagreement-penalized robustness.}
The disagreement-penalized score is
\begin{equation}
S_{\mathrm{UP}}(t)=
\bar{\rho}(t)-\lambda U_\rho(t),
\end{equation}
where $\lambda\geq0$ controls the penalty strength.
In our implementation, $P$ is the ensemble mean of the
per-member composite robustness scores, and therefore
$S_{\mathrm{UP}}=P$ at $\lambda=0$.
We evaluate $\lambda\in\{0,1,2,5\}$ to examine sensitivity
to the penalty strength. The values are fixed across test
conditions and are not tuned using shifted test data.
A window is flagged when $S_{\mathrm{UP}}<0$, using the
STL satisfaction boundary as a common threshold across
conditions and prediction horizons.

The penalty should not be interpreted as a formal probabilistic
confidence bound. It reduces the predicted robustness margin
in proportion to ensemble disagreement, but a lower score
does not, by itself, establish improved failure detection.

\paragraph{Failure prediction.}
We evaluate the ability of the disagreement and robustness
scores to rank future violations using AUROC and AUPRC.
The evaluated scores are $U_\rho$, $U_o$, $U_z$,
$-\bar{\rho}$, and $S_{\mathrm{UP}}$.
We consider both per-specification labels
$Y_{i,t}^{(K)}$ and the any-violation label $Y_t^{(K)}$,
as defined in the main paper.

\paragraph{Error association.}
To examine whether disagreement reflects prediction error,
we analyze the relationship between disagreement and absolute
STL robustness prediction error. In particular, we examine
the conditional mean error
$\mathbb{E}[|\rho-\hat{\rho}|\mid U]$
across disagreement buckets. If disagreement is informative
about prediction error, this error should generally increase
with disagreement. Pearson's correlation coefficient is
reported as a secondary descriptive diagnostic.

\paragraph{Selective prediction.}
We evaluate disagreement-based abstention by rejecting
windows whose STL disagreement exceeds a threshold:
\begin{equation}
U_\rho(t)>\tau.
\end{equation}
where $\tau$ is the empirical disagreement threshold selected to achieve
the specified coverage level.
We report the retained violation rate at coverage levels
of $100\%$, $90\%$, $80\%$, and $70\%$. This analysis
examines whether removing windows with high disagreement
reduces the violation rate among the retained predictions.

\subsection{Future STL Violation: Legacy Correlations and Horizon Illustration}

\begin{table}[t]
\centering
\caption{Correlation ($r$) between STL robustness disagreement and
future STL violation, and legacy single-seed window-level composite
violation rate (\%) by horizon, closed-loop. The Ant-v5 violation
rates refer to the predictive composite and are zero at all reported
horizons. The legacy single-seed runs used 3 episodes per condition ($n{=}27$ windows at $K{=}10$), whereas the seed-0 horizon analysis in Appendix~\ref{app:sup-primary} (Tab.~\ref{tab:horizon_auroc}) uses 10 episodes ($n{=}90$); the composite-label rates for noise $\sigma{=}0.15$ at $K{=}10$ ($37.0\%$ vs.\ $35.6\%$) therefore differ by sampling, not by definition.}
\label{tab:future_violation}
\begin{tabular}{lcccc}
\toprule
& \multicolumn{4}{c}{Horizon $K$} \\
\cmidrule(lr){2-5}
Condition & $K{=}1$ & $K{=}5$ & $K{=}10$ & $K{=}20$ \\
\midrule
\multicolumn{5}{l}{\textit{Ant-v5: violation rate (\%)}} \\
All conditions & 0.0 & 0.0 & 0.0 & 0.0 \\
\midrule
\multicolumn{5}{l}{\textit{HalfCheetah-v5: correlation ($r$)}} \\
baseline     & $-0.07$ & $-0.04$ & $-0.20$ & $-0.37$ \\
force 5N     & $-0.01$ & $-0.14$ & $-0.26$ & $-0.57$ \\
force 15N    & $-0.07$ & $+0.04$ & $-0.08$ & $-0.15$ \\
noise 0.05   & $-0.13$ & $-0.16$ & $-0.14$ & $-0.20$ \\
noise 0.15   & $-0.09$ & $+0.06$ & $+0.27$ & $+0.27$ \\
mass 0.8x    & $+0.03$ & $-0.06$ & $+0.08$ & $+0.00$ \\
mass 1.6x    & $-0.07$ & $-0.15$ & $-0.23$ & $-0.35$ \\
\midrule
\multicolumn{5}{l}{\textit{HalfCheetah-v5: violation rate (\%)}} \\
baseline     & 1.0 & 3.5 & 7.4 & 16.7 \\
force 5N     & 0.7 & 3.5 & 7.4 & 8.3 \\
force 15N    & 1.0 & 3.5 & 7.4 & 8.3 \\
noise 0.05   & 1.7 & 7.0 & 11.1 & 16.7 \\
noise 0.15   & 7.7 & 29.8 & 37.0 & 66.7 \\
mass 0.8x    & 0.7 & 1.8 & 3.7 & 0.0 \\
mass 1.6x    & 1.3 & 5.3 & 11.1 & 16.7 \\
\bottomrule
\end{tabular}
\end{table}

\begin{figure}[t]
\centering

\begin{tikzpicture}
\begin{axis}[
  width=0.48\columnwidth,
  height=5.5cm,
  xlabel={Condition},
  ylabel={Future violation rate (\%)},
  ymin=0, ymax=75,
  xtick={1,2,3,4,5,6,7},
  xticklabels={base, 5N, 15N, n.05, n.15, m.8, m.16},
  x tick label style={
    rotate=45,
    anchor=east,
    font=\scriptsize
  },
  y tick label style={font=\scriptsize},
  ybar,
  bar width=6pt,
  grid=major,
  grid style={gray!30},
  title={\textbf{HalfCheetah-v5: Future violations at $K=20$}},
  title style={font=\small},
]

\addplot[
  fill=blue!60
] coordinates {
  (1,16.7)
  (2,8.3)
  (3,8.3)
  (4,16.7)
  (5,66.7)
  (6,0.0)
  (7,16.7)
};

\end{axis}
\end{tikzpicture}
\hfill
\begin{tikzpicture}
\begin{axis}[
  width=0.48\columnwidth,
  height=5.5cm,
  xlabel={$r_{UV}$ (point-biserial correlation)},
  ylabel={Future violation rate (\%)},
  ymin=0, ymax=75,
  xmin=-0.65, xmax=0.35,
  grid=major,
  grid style={gray!30},
  title={\textbf{Disagreement--violation alignment is inconsistent}},
  title style={font=\small},
]

\addplot[
  only marks,
  mark=*,
  mark size=3pt
] coordinates {
  (-0.37,16.7) 
  (-0.57,8.3)  
  (-0.15,8.3)  
  (-0.20,16.7) 
  (0.27,66.7)  
  (0.00,0.0)   
  (-0.35,16.7) 
};

\node[
  font=\scriptsize,
  anchor=west
] at (axis cs:0.27,62)
{noise $\sigma=0.15$};

\draw[
  thin,
  dashed
] (axis cs:0.27,66.7) -- (axis cs:0.27,60);

\end{axis}
\end{tikzpicture}

\caption{
\textbf{Left:} Future violation rate for HalfCheetah-v5 at $K=20$
across representative distribution shifts. Severe observation noise
produces the highest violation rate (66.7\%). 
\textbf{Right:} Point-biserial correlation $r_{UV}$ between ensemble
STL robustness disagreement and future violation across the same conditions.
The correlations vary in both magnitude and direction; notably, the
condition with the highest violation rate has only a moderate positive
correlation ($r_{UV}=0.27$), while several lower-violation conditions
exhibit negative correlations. These condition-level results provide
no consistent evidence that ensemble STL robustness disagreement reliably identifies
future STL violations.
}

\label{fig:uncertainty_failure}
\end{figure}

Under the selected Ant-v5 specifications, none of the evaluated monitoring
windows produced a negative composite robustness score. This result does
not contradict the 40\% episode-level violation rate reported in
Appendix~\ref{app:sup-legacy}, because the predictive-window analysis uses
shorter $K$-step windows and excludes the recovery specification from the
composite, whereas the episode-level analysis evaluates full 500-step
trajectories using all six specifications. Moreover, horizon-dependent
analysis for Ant-v5 is limited because several shifted conditions terminate
before 20 steps, leaving insufficient data for reliable estimates at
$K{=}20$.

For HalfCheetah-v5, the fraction of windows containing at least one
violation increases with the prediction horizon, with the strongest effect
under observation noise (e.g., in this legacy single-seed analysis the rate
rises from 7.7\% at $K{=}1$ to 66.7\% at $K{=}20$ under noise
$\sigma{=}0.15$, Tab.~\ref{tab:future_violation}; the seed-0 horizon
analysis gives 10.2\% to 67.5\%, Tab.~\ref{tab:horizon_auroc}, and the
multi-seed aggregate is $64.2{\pm}24.4\%$ at $K{=}20$). Some individual
conditions exhibit moderate associations between ensemble STL robustness
disagreement and future violations, but this association varies in both
direction and magnitude across shifts and horizons, providing no
consistent evidence that $U_\rho$ distinguishes windows containing future
violations from those without them. Consistent with this, the main
analysis finds that the disagreement--violation correlation is weak and
that $U_\rho$ does not reliably indicate when predictions are trustworthy
(Section~\ref{sec:results}). A concrete illustration of optimistic
predictive bias arises under noise $\sigma{=}0.15$ at $K{=}20$, where the
predictive score stays at $+0.15$ while the retrospective score drops to
$+0.01$ and 66.7\% of windows contain violations in this legacy
single-seed slice, yet the disagreement--violation correlation is only
$+0.27$.

\subsection{Disagreement-Penalized Score}

We evaluate $S_{\mathrm{UP}}=\bar{\rho}-\lambda U_\rho$ for
$\lambda\in\{0,1,2,5\}$. Table~\ref{tab:lcb} reports the components for
the Ant-v5 force 15N condition and HalfCheetah-v5 noise $\sigma{=}0.15$ at
$K{=}20$, the latter chosen because it has the lowest mean robustness
among conditions with $K{=}20$ data. The same disagreement penalty
produces a similar relative reduction in both examples (32\% for Ant-v5
and 33\% for HalfCheetah-v5), but its absolute decision impact is greater
when the mean robustness has a smaller positive margin to the zero
boundary. In the HalfCheetah-v5 case, the score approaches zero without
crossing it. These examples demonstrate sensitivity of the reported
robustness margin to disagreement penalization, not an improvement in
safety-detection performance. Neither example crosses the violation
boundary.

\begin{table}[t]
\centering
\caption{$S_{\mathrm{UP}}$ components, $K{=}20$, closed-loop. Ant-v5
under force 15N; HalfCheetah-v5 under noise $\sigma{=}0.15$.}
\label{tab:lcb}
\begin{tabular}{lcccc}
\toprule
& $\bar{\rho}$ & $\lambda U_\rho$ & $S_{\mathrm{UP}}$ \\
\midrule
\multicolumn{4}{l}{\textit{Ant-v5 (force 15N)}} \\
$\lambda{=}0$ & $+0.72$ & $0.000$ & $+0.72$ \\
$\lambda{=}1$ & $+0.72$ & $0.045$ & $+0.67$ \\
$\lambda{=}2$ & $+0.72$ & $0.090$ & $+0.63$ \\
$\lambda{=}5$ & $+0.72$ & $0.226$ & $+0.49$ \\
\midrule
\multicolumn{4}{l}{\textit{HalfCheetah-v5 (noise $\sigma{=}0.15$)}} \\
$\lambda{=}0$ & $+0.15$ & $0.000$ & $+0.15$ \\
$\lambda{=}1$ & $+0.15$ & $0.010$ & $+0.14$ \\
$\lambda{=}2$ & $+0.15$ & $0.020$ & $+0.13$ \\
$\lambda{=}5$ & $+0.15$ & $0.050$ & $+0.10$ \\
\bottomrule
\end{tabular}
\end{table}

Although neither example crosses the violation boundary, a sufficiently
large penalty could push a small positive margin below zero. A limitation
of this construction is that the penalized score conflates model
disagreement with safety; when the baseline robustness is already close
to zero, this conflation has practical consequences.

\section{Supplementary Primary-Analysis Details}\label{app:sup-primary}

\subsection{Per-Specification Violation Details}

Two distinct operations are applied to the recovery property throughout
the evaluation. First, it is excluded from the predictive composite
robustness because its 50-step temporal obligation is longer than the
$K$-step prediction windows considered here. Second, when recovery
robustness is reported per specification, it is computed with pastified
online semantics on the available $K$-step trace. The resulting value
summarizes the available trace prefix rather than establishing whether the
original 50-step recovery obligation has been satisfied; the full
interpretation requires a sufficiently long trace ($K\geq50$). Although
recovery is excluded from the predictive composite at $K<50$, its
truncated robustness remains part of the per-specification any-violation
target $Y_t^{(K)}$ (Sec.~\ref{sec:stl}). The per-specification results below
use this truncated computation and should be interpreted accordingly.

\begin{table}[t]
\centering
\caption{Per-specification (binary, each spec alone) violation rate (\%)
at $K{=}10$, seed 0 (10 episodes). Recovery robustness is computed on the
truncated $K$-step window; the full 50-step interpretation is available
only at $K{\geq}50$. These per-spec rates use their own single-spec
targets and are not comparable to the composite-label rate in
Tab.~\ref{tab:horizon_auroc} or the any-violation rate in Tab.~\ref{tab:auroc}
(same seed-0 windows under noise 0.15: height 95.6\%, composite 35.6\%,
any-violation 95.6\%). Composite masks per-spec failures: Ant composite
0\% while \texttt{no\_falls}/recovery violate frequently; HalfCheetah
violations concentrate in \texttt{no\_falls}/recovery while
\texttt{stability\_pitch}/\texttt{contact\_safety} are near-constant.}
\label{tab:perspec}
\begin{tabular}{lcccc}
\toprule
 & \multicolumn{2}{c}{Ant-v5} & \multicolumn{2}{c}{HalfCheetah-v5} \\
Condition & no\_falls & recovery & no\_falls & recovery \\
\midrule
baseline & 88.9 & 88.9 & 90.0 & 64.4 \\
noise 0.15 & 100 & 100 & 95.6 & 50.0 \\
mass 1.6x & 79.2 & 75.0 & 98.9 & 76.7 \\
friction 0.5x & 100 & 100 & 95.6 & 66.7 \\
compound & 100 & 100 & 98.9 & 75.6 \\
\bottomrule
\end{tabular}
\end{table}

The Ant predictive composite has no observed negative-robustness
windows (0\% composite-label prevalence across all evaluated
conditions and seeds, $n{=}13$--$49$ windows per slice at $K{=}10$); the weighted mean is dominated by the large
positive \texttt{velocity\_tracking} contribution (${\approx}+3.25$),
which can offset negative height and recovery robustness. The three
reported rates measure different targets on the same windows and are not
interchangeable: per-specification binary rates (Tab.~\ref{tab:perspec};
e.g.\ height 95.6\% under noise 0.15, seed 0, $K{=}10$), the composite-label
rate (Tab.~\ref{tab:horizon_auroc}; 35.6\% for that slice, corresponding
to negative observed composite robustness), and the any-violation rate
(Tab.~\ref{tab:auroc}; 95.6\% for seed 0 and
$55.9{\pm}28.6\%$ across seeds, based on the union of the evaluated
per-specification labels).

HalfCheetah composite-score violations occur in $10.4{\pm}6.0\%$ of
windows at baseline and $26.3{\pm}7.7\%$ under noise 0.15 at $K{=}10$
($49.2{\pm}13.6\%$ at $K{=}20$), whereas the broader any-specification
violation target in Tab.~\ref{tab:auroc} occurs in $56{\pm}25\%$ and
$56{\pm}29\%$ of windows, respectively. Per-specification diagnostics
attribute failures almost entirely to \texttt{no\_falls} (85--99\%) and
recovery (50--77\%; truncated-window interpretation at $K{<}50$);
\texttt{stability\_pitch} is approximately 0\% (2.2\% under severe noise)
and \texttt{contact\_safety} is 0\%. Per-specification binary prediction
is therefore primary; composites are secondary indices. Ant
per-specification failure ranking, where AUROC is defined (baseline and
mass 1.6x; remaining conditions have 100\% per-specification violation
rates), gives $-P$ AUROC $0.75$--$0.86$ on \texttt{no\_falls} and $1.0$
on recovery, with $U_\rho$ partially informative ($0.73$--$0.83$).
Alternative representations on the same multi-seed any-violation target
($K{=}10$) give $U_o$/$U_z$ AUROC $0.45{\pm}0.06$/$0.54{\pm}0.06$ at
baseline, $0.51{\pm}0.11$/$0.48{\pm}0.01$ under noise 0.15, and
$0.45{\pm}0.09$/$0.44{\pm}0.04$ at mass 1.6x.

\subsection{Ant Any-Violation Target}

The Ant any-violation label is the union of the evaluated
per-specification window labels:
\begin{equation}
Y_t^{(K)}
=
\mathbf{1}\left[
\exists i\in\mathcal{I}_{\mathrm{eval}}^{e}(K):
Y_{i,t}^{(K)}=1
\right],
\end{equation}
where $\mathcal{I}_{\mathrm{eval}}^{e}(K)$ contains all
per-specification labels produced by the evaluation for environment
$e$ and horizon $K$, $i$ indexes those evaluated specifications, and
$\mathbf{1}[\cdot]$ is the indicator function. For Ant-v5, this includes the truncated recovery
label when $K<50$, even though recovery is excluded from the
predictive composite at those horizons. The any-violation target is
therefore not restricted to the five properties included in the
predictive composite. The recovery label represents the available
trace prefix and must not be interpreted as establishing whether the
complete 50-step recovery obligation has been satisfied. The results
use the same monitoring windows as the per-specification rates in
Tab.~\ref{tab:perspec}.

Across 8 evaluated conditions and 3 seeds (24 seed-condition slices at
$K{=}10$), any-violation prevalence is $79$--$100\%$. Twenty-one of the
24 slices are $100\%$ positive, so AUROC is undefined there because no
\emph{negative} class exists (not because no positives exist). The three
slices admitting both classes are listed in Tab.~\ref{tab:ant_any}. In
all three, $-P$ ranks violations near-perfectly while $U_\rho$ ranges from
near chance ($0.42$) to partially informative ($0.83$); with
$n_{\mathrm{neg}}$ as low as $2$, these slices are descriptive only.
These results provide per-specification evidence about the evaluated
safety properties, but they do not establish a broadly usable
Ant-v5 any-violation discrimination benchmark: the strong label
imbalance makes AUROC undefined in most seed-condition slices.
Ant therefore contributes per-specification violation rates
(Tab.~\ref{tab:perspec}), the three both-class slices in
Tab.~\ref{tab:ant_any}, and evidence that composite-level and
any-violation targets can exhibit very different prevalence. This
distinction is why the quantitative failure-prediction analysis in the
main text is centered on HalfCheetah-v5.

\begin{table}[t]
\centering
\caption{Ant-v5 any-violation target: the three seed-condition slices
($K{=}10$) whose labels admit both classes; all other evaluated slices
are $100\%$ positive (single class). AUROC is undefined for the remaining
slices. The any-violation label is the union of all evaluated
per-specification labels, including truncated recovery labels. Seeds are
PPO training seeds $\{0,42,123\}$.}
\label{tab:ant_any}
\small
\begin{tabular}{llccccc}
\toprule
Condition & seed & $n$ & Viol\% & $-P$ & $U_\rho$ & $-O_t$ \\
\midrule
baseline & 0 & 36 & 88.9 & 1.00 & 0.76 & 0.81 \\
mass 1.6x & 0 & 24 & 79.2 & 0.98 & 0.83 & 0.87 \\
friction 0.5x & 42 & 44 & 95.5 & 1.00 & 0.42 & 1.00 \\
\bottomrule
\end{tabular}
\end{table}

\subsection{Horizon Sensitivity (Severe Noise)}

Table~\ref{tab:horizon_auroc} and Fig.~\ref{fig:horizon_sensitivity} give
the full multi-metric numbers and trend underlying the main-text horizon
summary.

\pgfplotsset{compat=1.18}

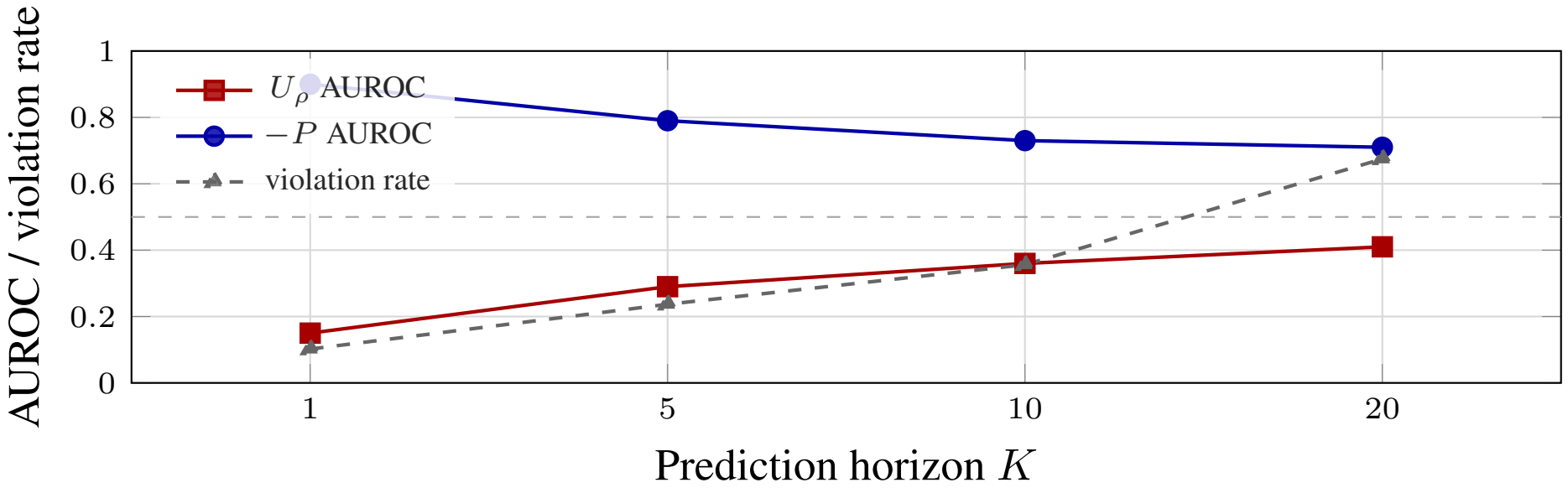
\begin{figure}[t]
\centering
\begin{tikzpicture}
\begin{axis}[
  width=0.92\columnwidth,
  height=4.2cm,
  xlabel={Prediction horizon $K$},
  ylabel={AUROC / violation rate},
  xmin=0.5, xmax=4.5,
  ymin=0.0, ymax=1.0,
  xtick={1,2,3,4},
  xticklabels={$1$,$5$,$10$,$20$},
  y tick label style={font=\scriptsize},
  x tick label style={font=\scriptsize},
  grid=major,
  grid style={gray!30},
  legend style={
    at={(0.02,0.98)},
    anchor=north west,
    font=\scriptsize,
    draw=none,
    fill=white,
    fill opacity=0.85
  },
]

\addplot[gray!70, dashed, thin, forget plot]
  coordinates {(0.5,0.5) (4.5,0.5)};

\addplot[
  mark=square*,
  mark size=2pt,
  thick,
  red!65!black
] coordinates {
  (1, 0.15)
  (2, 0.29)
  (3, 0.36)
  (4, 0.41)
};
\addlegendentry{$U_\rho$ AUROC}

\addplot[
  mark=*,
  mark size=2pt,
  thick,
  blue!65!black
] coordinates {
  (1, 0.90)
  (2, 0.79)
  (3, 0.73)
  (4, 0.71)
};
\addlegendentry{$-P$ AUROC}

\addplot[
  mark=triangle*,
  mark size=2pt,
  thick,
  dashed,
  gray!80!black
] coordinates {
  (1, 0.102)
  (2, 0.237)
  (3, 0.356)
  (4, 0.675)
};
\addlegendentry{violation rate}

\end{axis}
\end{tikzpicture}

\caption{
Horizon sensitivity under severe noise (HalfCheetah noise $\sigma{=}0.15$, seed 0; values from Tab.~\ref{tab:horizon_auroc}). Longer horizons expose more violations (10.2\% at $K{=}1$ to 67.5\% at $K{=}20$), yet $U_\rho$ AUROC climbs only from 0.15 toward 0.41 while $-P$ degrades gracefully from 0.90 to 0.71. Dashed gray line marks chance (0.5). Single-seed illustration of the pattern reported multi-seed at $K{=}10$/$20$ in Sec.~\ref{sec:extension} (Tab.~\ref{tab:future_violation} reports 66.7\% at $K{=}20$ on the legacy single-seed windowing).
}
\label{fig:horizon_sensitivity}
\end{figure}

\begin{table}[htbp]
\centering
\caption{Horizon sensitivity under severe noise (HalfCheetah noise
$\sigma{=}0.15$, seed 0). Viol\% is the composite-label rate (fraction of
windows with negative observed composite), not the per-specification or
any-violation rates of Tab.~\ref{tab:perspec}/Tab.~\ref{tab:auroc}; on the
same windows the any-violation rate is higher (e.g.\ 95.6\% at $K{=}10$).
$-P$ ranks failures across horizons while $U_\rho$ remains at or below
chance. Baseline shows the same pattern at lower prevalence.}
\label{tab:horizon_auroc}
\begin{tabular}{lcccccc}
\toprule
$K$ & $n$ & Viol\% & AUROC($U_\rho$) & AUROC($-P$) & AUPRC($U_\rho$) & AUPRC($-P$) \\
\midrule
1 & 990 & 10.2 & 0.15 & 0.90 & 0.06 & 0.74 \\
5 & 190 & 23.7 & 0.29 & 0.79 & 0.17 & 0.67 \\
10 & 90 & 35.6 & 0.36 & 0.73 & 0.28 & 0.63 \\
20 & 40 & 67.5 & 0.41 & 0.71 & 0.64 & 0.82 \\
\bottomrule
\end{tabular}
\end{table}

\subsection{Current-State Baseline and Safe-at-Start Analysis}

The instantaneous composite is
\[
O_t=\frac{1}{|\mathcal{I}_{\mathrm{pt}}|}
\sum_{i\in\mathcal{I}_{\mathrm{pt}}}\rho_i(x_t),
\]
where $x_t=q(o_t)$ is the safety-variable vector extracted from the current
observation, $\rho_i(x_t)$ is the pointwise robustness of specification $i$,
and $\mathcal{I}_{\mathrm{pt}}\subseteq\mathcal{I}_e(K)$ contains the
pointwise-applicable specifications and excludes response properties such
as recovery, which are not defined at a single instant. It evaluates the
same safety variables, thresholds, and predicates as the predictive
monitor, but without a future rollout. In the evaluation implementation,
both $O_t$ and the predictive composite $P$ are computed as unweighted
means of the applicable per-specification robustness values. The weighted
expression in Sec.~\ref{sec:stl} is the general formulation; the reported
experiments use equal aggregation weights. Thus, the two scores use the
same aggregation principle and pointwise-applicable predicates, and differ
in their temporal evaluation: $O_t$ reads the current observation, whereas
$P$ evaluates a policy-conditioned world-model rollout.

Table~\ref{tab:current_state} compares their failure-ranking performance
and reports the \emph{pointwise-safe-at-start} analysis at $K{=}10$
(multi-seed; pointwise-safe-at-start means that all pointwise predicates
are satisfied at the query time). Across all windows, $-O_t$ achieves
AUROCs close to those of $-P$, with differences of at most $0.03$ across
the reported conditions. Among windows satisfying all pointwise predicates
at their start, future violations remain frequent ($30$--$45\%$,
depending on the condition). However, $-O_t$ and $-P$ again exhibit
similar ranking performance (safe-at-start AUROC differences at most
$0.02$), whereas $U_\rho$ provides weaker and less consistent
discrimination ($0.30$--$0.67$). Thus, future violations can occur after
an initially predicate-satisfied state, but these results do not establish
a substantial ranking advantage from the world-model rollout over
instantaneous robustness; ensemble disagreement contributes little
additional information about future violations in this setting. Ant-v5
has very few pointwise-safe-at-start windows: $3.7{\pm}5.2\%$ (baseline),
$0\%$ (force 15N), $1.9{\pm}2.6\%$ (noise 0.15), and $8.3{\pm}11.8\%$
(mass 1.6x; driven by a $25\%$ seed-0 slice), i.e.\ its any-violation
windows are almost all already unsafe at the monitor's query time.

\begin{table*}[t]
\centering
\caption{Current-state baseline and safe-at-start analysis,
HalfCheetah-v5, $K{=}10$, mean$\pm$std over 3 seeds (any-violation
target). $-O_t$: instantaneous robustness at time $t$ (no prediction).
Safe: all pointwise predicates satisfied at $t$; ``safe viol.\%'' is the
future ($t{+}1$..$t{+}K$) any-violation rate within that subset. SAFE
columns: AUROC restricted to safe-at-start windows with the future-only
label.}
\label{tab:current_state}
\footnotesize
\setlength{\tabcolsep}{3.5pt}
\begin{tabular}{lcccccccc}
\toprule
 & \multicolumn{3}{c}{All windows (AUROC)} & \multicolumn{5}{c}{Safe-at-start} \\
\cmidrule(lr){2-4}\cmidrule(lr){5-9}
Condition & $-O_t$ & $-P$ & $U_\rho$ & Safe\% & viol\% & $-O_t$ & $-P$ & $U_\rho$ \\
\midrule
baseline & $0.90{\pm}0.09$ & $0.92{\pm}0.08$ & $0.50{\pm}0.08$ & $63{\pm}12$ & $36{\pm}31$ & $0.90{\pm}0.09$ & $0.90{\pm}0.10$ & $0.38{\pm}0.11$ \\
noise 0.05 & $0.79{\pm}0.26$ & $0.81{\pm}0.25$ & $0.50{\pm}0.07$ & $63{\pm}11$ & $38{\pm}38$ & $0.76{\pm}0.29$ & $0.76{\pm}0.30$ & $0.38{\pm}0.21$ \\
noise 0.15 & $0.97{\pm}0.01$ & $0.98{\pm}0.00$ & $0.56{\pm}0.18$ & $65{\pm}7$ & $36{\pm}40$ & $0.93{\pm}0.04$ & $0.95{\pm}0.05$ & $0.52{\pm}0.24$ \\
mass 0.8x & $0.93{\pm}0.07$ & $0.96{\pm}0.05$ & $0.47{\pm}0.09$ & $63{\pm}10$ & $32{\pm}28$ & $0.92{\pm}0.11$ & $0.91{\pm}0.12$ & $0.36{\pm}0.03$ \\
mass 1.6x & $0.92{\pm}0.02$ & $0.94{\pm}0.01$ & $0.47{\pm}0.03$ & $60{\pm}10$ & $45{\pm}40$ & $0.86{\pm}0.06$ & $0.87{\pm}0.06$ & $0.62{\pm}0.12$ \\
friction 0.5x & $0.86{\pm}0.15$ & $0.88{\pm}0.14$ & $0.58{\pm}0.13$ & $63{\pm}12$ & $38{\pm}39$ & $0.83{\pm}0.16$ & $0.83{\pm}0.16$ & $0.67{\pm}0.08$ \\
compound & $0.96{\pm}0.01$ & $0.97{\pm}0.01$ & $0.45{\pm}0.02$ & $61{\pm}9$ & $45{\pm}38$ & $0.94{\pm}0.01$ & $0.95{\pm}0.02$ & $0.30{\pm}0.20$ \\
force 5N & $0.95{\pm}0.04$ & $0.96{\pm}0.04$ & $0.51{\pm}0.07$ & $66{\pm}10$ & $31{\pm}27$ & $0.93{\pm}0.07$ & $0.94{\pm}0.07$ & $0.45{\pm}0.02$ \\
force 15N & $0.95{\pm}0.03$ & $0.95{\pm}0.03$ & $0.52{\pm}0.07$ & $70{\pm}13$ & $30{\pm}28$ & $0.91{\pm}0.08$ & $0.91{\pm}0.08$ & $0.47{\pm}0.10$ \\
\bottomrule
\end{tabular}
\end{table*}

\subsection{Height-Threshold Sensitivity}

Table~\ref{tab:height_sens} varies the \texttt{no\_falls} height threshold
around the default $h_{\min}{=}0.10$ (HalfCheetah, $K{=}10$, multi-seed).
The default was selected manually as a rounded safety margin below the
observed 1st percentile of nominal body height ($0.116$\,m); the tested
range $\{0.08,0.10,0.12\}$ brackets this default, with the upper end
slightly exceeding the nominal 1st percentile but remaining far below
nominal running heights, so all three thresholds still mark the vast
majority of nominal steps as safe. Any-violation prevalence and $-P$
AUROC shift only marginally, and $-P{\gg}U_\rho$ holds at every tested
threshold. The observed ranking is therefore robust across the three
tested height thresholds. This check addresses threshold choice only;
unit-normalized composite variants (rescaling heterogeneous robustness
quantities before averaging) remain untested.

\begin{table}[t]
\centering
\caption{Height-threshold sensitivity (\texttt{no\_falls} $h_{\min}$),
HalfCheetah-v5, $K{=}10$, mean$\pm$std over 3 seeds, any-violation target.}
\label{tab:height_sens}
\begin{tabular}{lcccccc}
\toprule
 & \multicolumn{3}{c}{baseline} & \multicolumn{3}{c}{noise 0.15} \\
\cmidrule(lr){2-4}\cmidrule(lr){5-7}
$h_{\min}$ & viol\% & $-P$ & $U_\rho$ & viol\% & $-P$ & $U_\rho$ \\
\midrule
0.08 & $54.8{\pm}23.7$ & $0.90{\pm}0.10$ & $0.51{\pm}0.09$ & $55.9{\pm}28.6$ & $0.98{\pm}0.00$ & $0.56{\pm}0.18$ \\
0.10 & $56.3{\pm}25.0$ & $0.92{\pm}0.08$ & $0.50{\pm}0.08$ & $55.9{\pm}28.6$ & $0.98{\pm}0.00$ & $0.56{\pm}0.18$ \\
0.12 & $57.0{\pm}26.0$ & $0.94{\pm}0.05$ & $0.48{\pm}0.05$ & $56.3{\pm}28.3$ & $0.99{\pm}0.00$ & $0.56{\pm}0.18$ \\
\bottomrule
\end{tabular}
\end{table}

\subsection{Ensemble-Size Sensitivity (Full Detail)}

\textbf{$M$-ablation over $\{1,2,3,5,10\}$.}
$M{=}1$ gives $U{\equiv}0$ (AUROC N/A). The magnitude of $U_\rho$
stabilizes for $M{\ge}3$ (HC noise $K{=}10$ s0:
$0.0064{\to}0.0118{\to}0.0107{\to}0.0113$), while increasing ensemble
size does not consistently improve failure discrimination:
\begin{center}
\begin{tabular}{lccccc}
\toprule
HC $K{=}10$ s0 & $M{=}2$ & $M{=}3$ & $M{=}5$ & $M{=}10$ \\
\midrule
baseline AUROC & 0.15 & 0.69 & 0.65 & 0.51 \\
noise 0.15 AUROC & 0.36 & 0.36 & 0.34 & 0.33 \\
mass 1.6x AUROC & 0.18 & 0.72 & 0.66 & 0.58 \\
\bottomrule
\end{tabular}
\end{center}
At $K{=}20$ noise 0.15: AUROC
$0.38{\to}0.41{\to}0.46{\to}0.45$ for $M{=}2,3,5,10$ (near chance
throughout). $U_o$ increases with $M$ in this ablation
($0.19{\to}0.29{\to}0.36{\to}0.39$ at noise $K{=}10$), reflecting the
additional ensemble members included in the dispersion estimate, yet
remains non-predictive. Ant $U_\rho$ stabilizes ${\approx}0.04$ for
$M{\ge}3$ with sign-flipping $r$; composite $Y{=}0$ so AUROC is N/A at
all $M$.

\subsection{Noise-Model and Shift Ablation (Full Detail)}

\textbf{Noise-model and shift ablation.}
In a separate noise-injection ablation, observation noise is injected
into each predicted observation at every rollout step, using the
corresponding deployment-time noise setting. This differs from the
standard rollout, in which subsequent predicted observations are
generated by the learned dynamics model without reapplying the
deployment-time observation-noise process. Under severe noise (noise
0.15, $K{=}10$), the ablation has little effect (AUROC
$0.36{\to}0.33$, bias $+0.12{\to}+0.13$), whereas under moderate noise
(noise 0.05) it has a more visible effect (AUROC $0.26{\to}0.38$).

\subsection{Disagreement-Based Reliability and Selective Prediction (Full Numbers)}

\textbf{Reliability and selective prediction (multi-seed).}
Reliability buckets are non-monotonic: multi-seed mean
$|\rho{-}\hat\rho|$ per $U_\rho$ quintile at HC noise 0.15 $K{=}20$ is
$0.24,0.36,0.30,0.28,0.27$ (std up to $0.16$). On the composite target,
the retained violation rate changes little as coverage decreases across
seeds: at $K{=}20$ retained violation rates are
$49.2{\pm}13.6 / 47.2{\pm}17.1 / 44.8{\pm}19.2 / 47.6{\pm}19.0\%$
at 100/90/80/70\% coverage ($K{=}10$:
$26.3{\pm}7.7 / 27.2{\pm}9.6 / 29.2{\pm}10.8 / 29.6{\pm}10.5\%$).
On the any-violation target the pattern is the same (Tab.~\ref{tab:sup_any}).
In the evaluated severe-noise setting, $U$-based abstention did not reduce
retained violation risk on either target.

\begin{table}[t]
\centering
\caption{Disagreement-penalized score
$S_{\mathrm{UP}}{=}P{-}\lambda U_\rho$ and risk--coverage on the
HalfCheetah any-violation target (noise $\sigma{=}0.15$,
mean$\pm$std over 3 seeds). Flag rate is the fraction of windows with
$S_{\mathrm{UP}}<0$ (the fixed STL satisfaction boundary, applied
uniformly across conditions and horizons). Penalization is flat
($\Delta$AUROC ${\leq}0.01$) while flag rate rises; abstention shows
little change across coverage levels.}
\label{tab:sup_any}
\small
\begin{tabular}{lcc}
\toprule
$\lambda$ & AUROC($-S_{\mathrm{UP}}$) & Flag rate \% \\
\midrule
\multicolumn{3}{l}{\textit{$K{=}10$}} \\
0 & $0.983{\pm}0.003$ & $5.2{\pm}1.0$ \\
1 & $0.983{\pm}0.002$ & $5.2{\pm}1.0$ \\
2 & $0.986{\pm}0.004$ & $5.9{\pm}1.0$ \\
5 & $0.989{\pm}0.004$ & $8.1{\pm}3.8$ \\
\midrule
\multicolumn{3}{l}{\textit{$K{=}20$}} \\
0 & $0.989{\pm}0.016$ & $3.3{\pm}2.4$ \\
1 & $0.990{\pm}0.015$ & $3.3{\pm}2.4$ \\
2 & $0.992{\pm}0.011$ & $4.2{\pm}3.1$ \\
5 & $0.997{\pm}0.005$ & $5.8{\pm}5.1$ \\
\midrule
\multicolumn{3}{l}{\textit{Risk--coverage (retained viol.\ \%)}} \\
 & $K{=}10$ & $K{=}20$ \\
100\% cov. & $55.9{\pm}28.6$ & $64.2{\pm}24.4$ \\
90\% cov. & $54.7{\pm}29.9$ & $62.0{\pm}26.5$ \\
80\% cov. & $54.6{\pm}29.9$ & $60.4{\pm}28.2$ \\
70\% cov. & $53.4{\pm}29.8$ & $60.7{\pm}27.8$ \\
\bottomrule
\end{tabular}
\end{table}

\end{document}